\documentclass{ieeeaccess}
\usepackage{cite}
\usepackage{amsmath,amssymb,amsfonts}
\usepackage{algorithmic}
\usepackage{graphicx}
\usepackage{textcomp}
\usepackage{amsmath,amsfonts}
\usepackage{algorithmic}
\usepackage{algorithm}
\usepackage{array}
\usepackage[caption=false,font=normalsize,labelfont=sf,textfont=sf]{subfig}
\usepackage{textcomp}
\usepackage{stfloats}
\usepackage{url}
\usepackage{verbatim}
\usepackage{graphicx}
\usepackage{cite}
\usepackage{enumitem}
\usepackage{adjustbox}
\usepackage{booktabs}
 \usepackage{caption} 
\usepackage{multirow}
\usepackage[table]{xcolor}
\usepackage{bm}
\usepackage[pagebackref,breaklinks,colorlinks,allcolors=blue]{hyperref}

\def\BibTeX{{\rm B\kern-.05em{\sc i\kern-.025em b}\kern-.08em
    T\kern-.1667em\lower.7ex\hbox{E}\kern-.125emX}}
\begin{document}
\history{Date of publication xxxx 00, 0000, date of current version xxxx 00, 0000.}
\doi{10.1109/ACCESS.2017.DOI}

\title{Face2Parts: Exploring Coarse-to-Fine Inter-Regional Facial Dependencies for Generalized Deepfake Detection}

\author{\uppercase{Kutub Uddin}\authorrefmark{1}, 
\uppercase{Nusrat Tasnim\authorrefmark{1}, and Byung Tae Oh}.\authorrefmark{1},
\IEEEmembership{Member, IEEE}}

\address[2]{School of Electronics and Information Engineering, Korea Aerospace University, Goyang, South Korea (e-mail: kutub@kau.ac.kr, tasnim.nishu70@kau.kr, byungoh@kau.ac.kr)}
\tfootnote{This work was supported by the IITP(Institute of Information \& Communications Technology Planning \& Evaluation)-ICAN(ICT Challenge and Advanced Network of HRD) grant funded by the Korea government(Ministry of Science and ICT)(IITP-2026-RS-2024-00437857) and by the 2025 Korea Aerospace University AIRBUS Foundation Research Grant.}

\markboth
{Author \headeretal: Preparation of Papers for IEEE TRANSACTIONS and JOURNALS}
{Author \headeretal: Preparation of Papers for IEEE TRANSACTIONS and JOURNALS}

\corresp{Corresponding author: Kutub Uddin (e-mail: kutub@kau.ac.kr).}

\begin{abstract}
Multimedia data, particularly images and videos, is integral to various applications, including surveillance, visual interaction, biometrics, evidence gathering, and advertising. However, amateur or skilled counterfeiters can simulate them to create deepfakes, often for slanderous motives. To address this challenge, several forensic methods have been developed to ensure the authenticity of the content.
The effectiveness of these methods depends on their focus, with challenges arising from the diverse nature of manipulations. In this article, we analyze existing forensic methods and observe that each method has unique strengths in detecting deepfake traces by focusing on specific facial regions, such as the frame, face, lips, eyes, or nose.
Considering these insights, we propose a novel hybrid approach called Face2Parts based on hierarchical feature representation ($HFR$) that takes advantage of coarse-to-fine information to improve deepfake detection. The proposed method involves extracting features from the frame, face, and key facial regions (i.e., lips, eyes, and nose) separately to explore the coarse-to-fine relationships. This approach enables us to capture inter-dependencies among facial regions using a channel-attention mechanism and deep triplet learning.
We evaluated the proposed method on benchmark deepfake datasets in both intra-, inter-dataset, and inter-manipulation settings. The proposed method achieves an average AUC of 98.42\% on FF++, 79.80\% on CDF1, 85.34\% on CDF2, 89.41\% on DFD, 84.07\% on DFDC, 95.62\% on DTIM, 80.76\% on PDD, and 100\% on WLDR, respectively. The results demonstrate that our approach generalizes effectively and achieves promising performance to outperform the existing methods.
\end{abstract}


\titlepgskip=-15pt

\maketitle

\section{Introduction}
\label{sec:introduction}
\IEEEPARstart{M}{odern} cameras have become more affordable and user-friendly, allowing anyone, regardless of technical expertise, to capture images and videos. As a result, they are now widely used in various fields, including visual communication, biometrics, evidence collection, surveillance systems, and election campaigns~\cite{suryadevara2021comprehensive}. Additionally, people frequently share and post images and videos with family and colleagues on social media platforms like Facebook, LinkedIn, and Instagram. However, these images and videos can be altered or generated ~\cite{seow2022comprehensive} by both amateur and professional counterfeiters using editing software or artificial intelligence (AI) techniques~\cite{verdoliva2020media} for unethical purposes~\cite{tasnim2025ai, tasnim2026grex}. AI technologies, particularly generative adversarial networks (GANs)~\cite{seow2022comprehensive, uddin2019anti}, are highly adept at creating realistic deepfakes, capable of deceiving even the human eye.\\
According to Surfshark~\cite{surfshark2024deepfake}, there were 22 reported deepfake incidents involving globally recognized individuals from 2017 to 2022. Remarkably, this number nearly doubled in just one year, reaching 42 cases in 2023. The trend escalated rapidly, with 150 cases in 2024 and 179 in 2025, marking an approximate increase of 257\% and 326\% compared to 2022, respectively. These significant threats to multimedia content, especially images and videos, have made deepfakes increasingly effective at deceiving human perception. Despite their impressive capabilities, GAN-generated deepfakes leave behind subtle artifacts that can reveal their synthetic nature~\cite{verdoliva2020media, uddin2025guard} to distinguish from authentic content.\\
\begin{figure}[!t]
    \centering
    \subfloat[\label{fig:ana_1}]{{\includegraphics[width=0.48\textwidth]{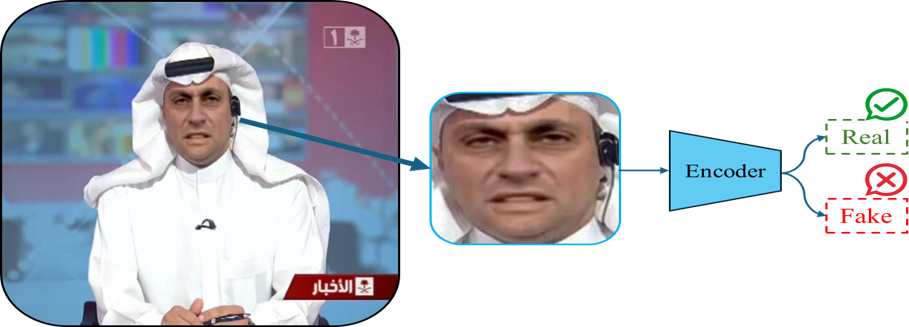}}} \\ \vspace{-10pt}
    \subfloat[\label{fig:ana_2}]{{\includegraphics[width=0.48\textwidth]{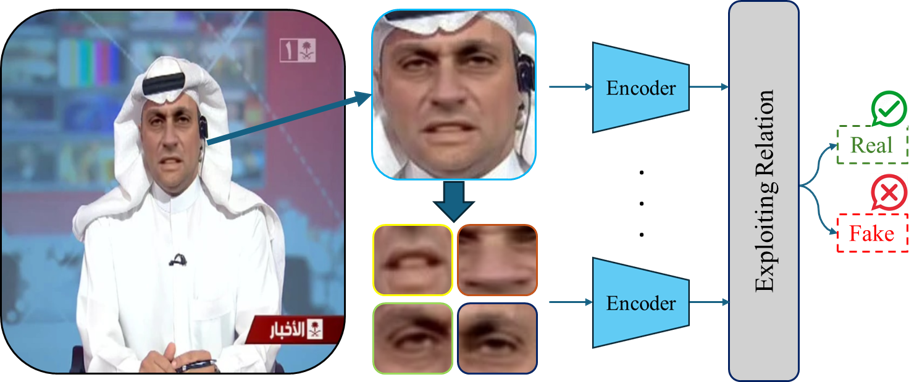}}} \vspace{-5pt}
    \caption{Key concepts of the proposed method: (a) The conventional deepfake detection, which relies solely on facial artifact analysis. (c) In contrast, the proposed method, which leverages hierarchical information across multiple levels, coarse (frame), medium (face), and fine-grained (lips, left eye, right eye, and nose), aims to improve deepfake detection performance.}
    \vspace{-15pt}
    \label{fig:intro}
\end{figure}
Several methods~\cite{rana2022deepfake} have been proposed to detect unnatural patterns in deepfakes. However, detecting deepfakes is highly dependent on the type of manipulations involved~\cite{rossler2019faceforensics++}, requiring specialized approaches for each manipulation. For example, a method trained on a dataset designed for face-swapping may not generalize well to other types, such as expression alteration. This limitation arises because most state-of-the-art (SoTA) methods rely on analyzing the face region alone to identify deepfake artifacts.\\
While a few methods~\cite{liu2024robust, hariprasad2022boundary} have attempted to identify common properties across different types of deepfakes to enable generalization, their effectiveness is often constrained by specific assumptions. It exhibits less performance when applied to diverse manipulations~\cite{nadimpalli2022improving}. Some methods are considered anti-forensic attacks~\cite{uddin2023robust, farooq2025transferable, uddin2021analysis} and countermeasures~\cite{uddin2024counter, uddin2025sheild} to provide robust deepfake detection. This highlights the need for more effective, generalized, and robust approaches to address the evolving challenges of deepfake detection. To address these gaps, we propose a novel hybrid approach based on hierarchical feature representation ($HFR$) as illustrated in Figure~\ref{fig:intro} to enhance generalization.\\
Unlike conventional methods~\cite{rana2022deepfake, liu2024robust} that primarily focus on the face region alone (Figure~\ref{fig:intro} (a)), the proposed approach extracts multi-level features (Figure~\ref{fig:intro} (b)), capturing coarse-, medium-, and fine-grained information. The proposed framework consists of three hierarchical levels of feature extraction: (a) level-1 (frame), which captures global features from the entire frame and provides coarse-grained information; (b) level-2 (face), which focuses on the cropped face region to capture medium-grained information; and (c) level-3 (left eye, right eye, lips, and nose), which extracts fine-grained information. This hierarchical approach ensures comprehensive detection by leveraging both global and local contexts, thereby significantly improving overall performance across unseen datasets.\\
The key insights of the article are summarized as follows:
\begin{itemize}[noitemsep,topsep=0pt]
\item We analyze SoTA methods and identify their unique strengths in detecting deepfake traces by focusing on specific facial regions such as the frame, face, lips, eyes, or nose, depending on the type of manipulation.
\item To enhance generalization, we propose a novel hybrid approach based on $HFR$ that extracts features from the frame, face, and key facial regions (left eye, right eye, lips, and nose), while capturing their inter-dependencies through a channel-attention mechanism within a deep triplet learning framework.
\item We demonstrate the effectiveness of the proposed $HFR$ through extensive evaluations on benchmark datasets, achieving promising performance in intra-dataset, inter-dataset, and inter-manipulation scenarios.
\end{itemize}

\section{Related Study}
\label{sec:related}
This section presents a detailed review of SoTA deepfake detection methods, grouped into conventional, deep learning-based, and hybrid approaches as shown in Figure~\ref{fig:rel_full}.
\vspace{-5pt}
\subsection{Conventional Deepfake Detection}
\begin{figure}[!t]
    \centering
    \subfloat[\label{fig:rel_a}]{{\includegraphics[width=0.9\columnwidth]{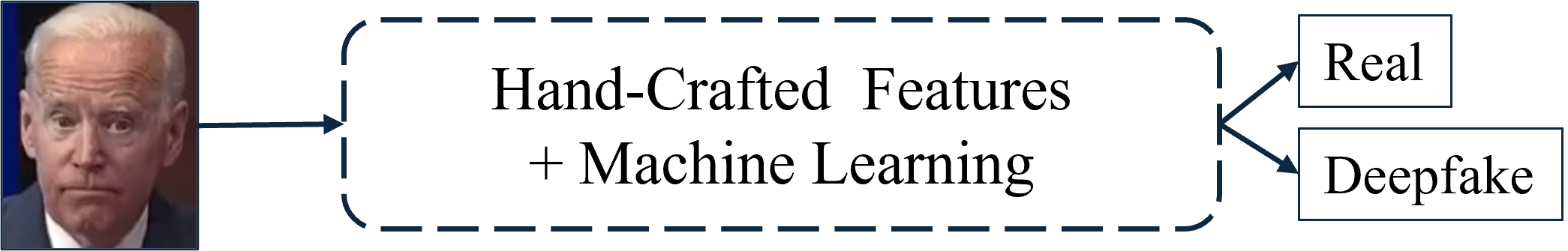}}} \\ \vspace{-8pt}
    \subfloat[\label{fig:rel_b}]{{\includegraphics[width=0.9\columnwidth]{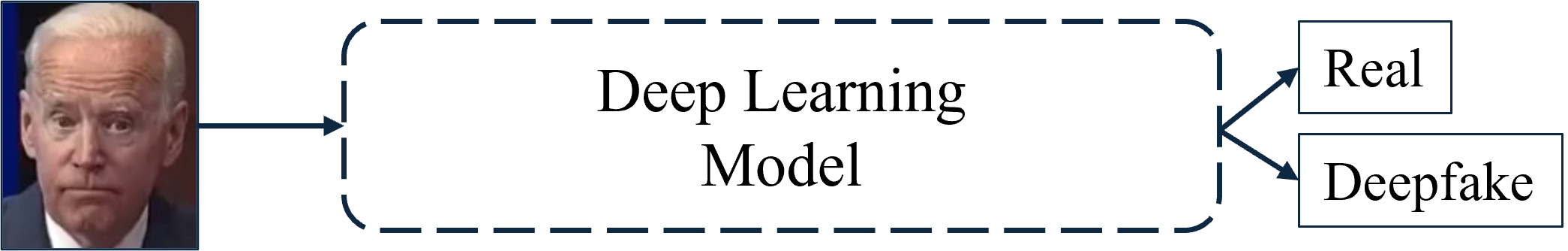}}} \\ \vspace{-8pt}
    \subfloat[\label{fig:rel_c}]{{\includegraphics[width=0.9\columnwidth]{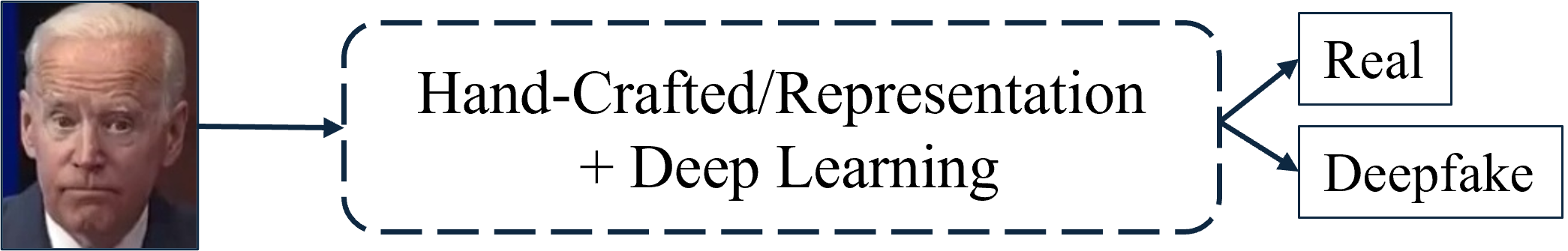}}}  \vspace{-5pt}
    \caption{Categories of deepfake detection methods: (a) conventional methods, which rely on handcrafted features, (b) deep learning-based methods that automatically learn discriminative representations from data, and (c) hybrid approaches that combine traditional cues with deep learning models to improve detection performance.}
    \vspace{-10pt}
    \label{fig:rel_full}
\end{figure}

Conventional approaches primarily rely on hand-crafted patterns, such as pixel correlations, facial asymmetries, or textures, to detect deepfakes using traditional machine learning techniques as depicted in Figure~\ref{fig:rel_full} (a).\\
For instance, Matern et al.~\cite{matern2019exploiting} utilized correlation features between the left and right eyes for deepfake detection. However, this method is fully dependent on color information from the eyes and does not account for other facial regions.
Guarnera et al.~\cite{guarnera2020fighting} explored pixel correlations to extract convolutional traces, employing machine learning techniques for classification. However, methods focusing solely on pixel relationships are less effective on more realistic and complex deepfakes. 
Agarwal et al. ~\cite{agarwal2019protecting} analyzed deepfakes of world leaders and extracted 20 facial and head pose features for detection. A similar approach was introduced by Yang~\cite{yang2019exposing}, where 3D head poses derived from landmarks of the face were used. This limits the method to specific, well-known subjects, leaving its applicability to the general public unexplored. 
Bonomi et al.~\cite{bonomi2021dynamic} examined both spatial and temporal textures in videos, leveraging local features. Despite being computationally efficient and simpler, conventional methods often suffer performance issues and are limited to specific datasets, features, and configurations.\\

\subsection{Deep Learning-based Deepfake Detection}
The limitations of conventional approaches lead to the development of deep learning-based methods, as depicted in Figure~\ref{fig:rel_full} (b), which consistently provide better performance. Several methods~\cite{passos2024review} have been introduced to detect deepfakes using deep learning. Afchar et al.~\cite{afchar2018mesonet} considered mesoscopic properties of images and designed two deep learning models, Meso-4 and MesoInception-4, for deepfake detection. These models showed promising performance on Deepfakes~\cite{garrido2014automatic} and Face2Face~\cite{thies2016face2face} datasets. 
Pan et al.~\cite{pan2020deepfake} directly implemented Xception~\cite{chollet2017xception} for deepfake detection on FaceForensic++ (FF++~\cite{rossler2019faceforensics++}) datasets and achieved promising performance.
Li et al.~\cite{li2020celeb} introduced a novel deepfake dataset, named Celeb-DF and evaluated the performance using Meso4~\cite{afchar2018mesonet}, Xception~\cite{chollet2017xception}, and more deep learning models.
Wang et al.~\cite{wang2020cnn} studied different deepfake generation methods and summarized that ProGAN could be used to train ResNet50~\cite{he2016deep} to generalize others, such as StyleGAN, BigGAN, CycleGAN, StarGAN, and more.
Yan et al.~\cite{yan2023deepfakebench} provided a comprehensive study on deepfake detection and evaluated the performance of different deepfake datasets using EfficientNet-B4~\cite{tan2019efficientnet}. Chen et al.~\cite{chen2022self} used GANs to generate augmented samples from different facial regions, thereby improving generalization across datasets. A similar approach was proposed by Yan et al.~\cite{yan2024transcending}, where augmentation was performed in the latent space. Steven et al.~\cite{schwarcz2021finding} introduced a patch-based method that employed multiple branches of the Xception model and made predictions using average pooling.
However, these models~\cite{afchar2018mesonet, chollet2017xception, he2016deep} often show less performance in different manipulations and lack generalization to unseen datasets. 

\subsection{Hybrid Deepfake Detection}
Hybrid methods incorporate both feature extraction or representation and deep learning for detection, as visualized in Figure~\ref{fig:rel_full} (c). 
Nataraj et al.~\cite{nataraj2019detecting} explored the co-occurrence matrix for red (R), green (G), and blue (B) channels to identify inconsistencies between real and deepfake samples. Barni et al.~\cite{barni2020cnn} extended the idea of co-occurrence matrix from Nataraj~\cite{nataraj2019detecting} and computed the cross-band co-occurrence matrix in cross-channel, such as  RG, RB, and GB, to improve overall performance. Then they applied a deep learning model for deepfake detection.
Rafique et al.~\cite{rafique2021deepfake} analyzed the error level and then applied deep learning for classification.
Zhang et al.~\cite{zhang2019detecting} studied the up- and down-sampling properties of deepfake images and extracted spectral features to distinguish them from real images.
Some methods~\cite{zhao2022self} have combined representation learning and deep learning to achieve SoTA performance. 
For example, Zhao et al.~\cite{zhao2022self} adopted a pre-trained model to generate embeddings in a feature space, then applied a deep learning model for classification. Although hybrid methods demonstrate promising performance, there is still room for improvement in terms of performance and generalization for practical applications.

\begin{figure}[!b]
\vspace{-18pt}
    \centering
    \subfloat[\label{fig:ana_1}]{{\includegraphics[width=0.5\textwidth]{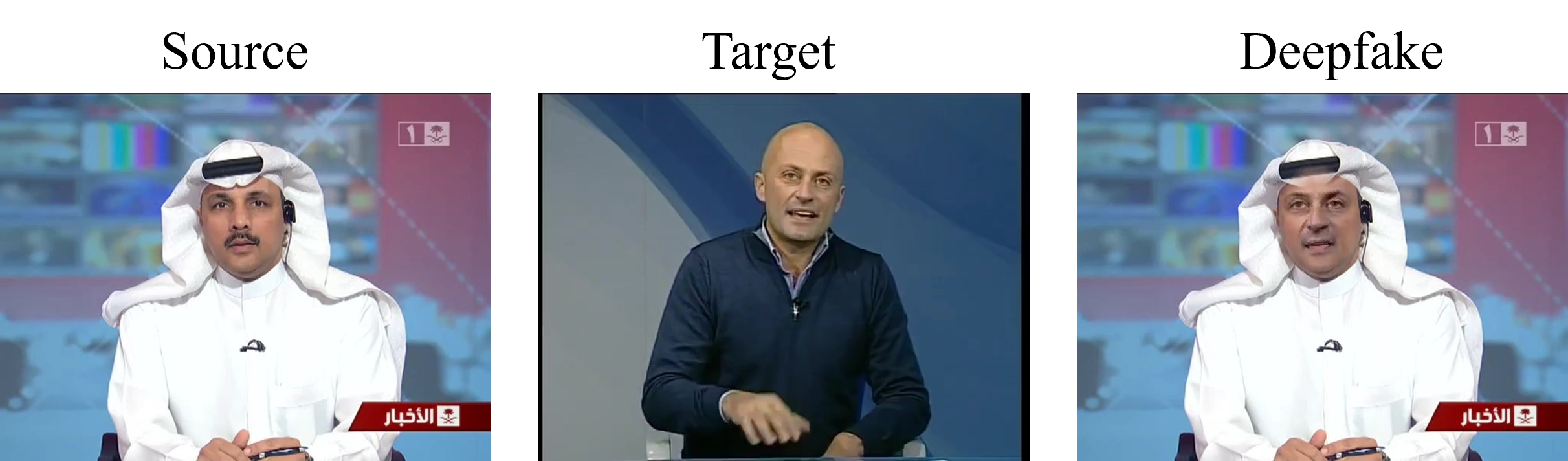}}} \\ \vspace{-10pt}
    \subfloat[\label{fig:ana_2}]{{\includegraphics[width=0.5\textwidth]{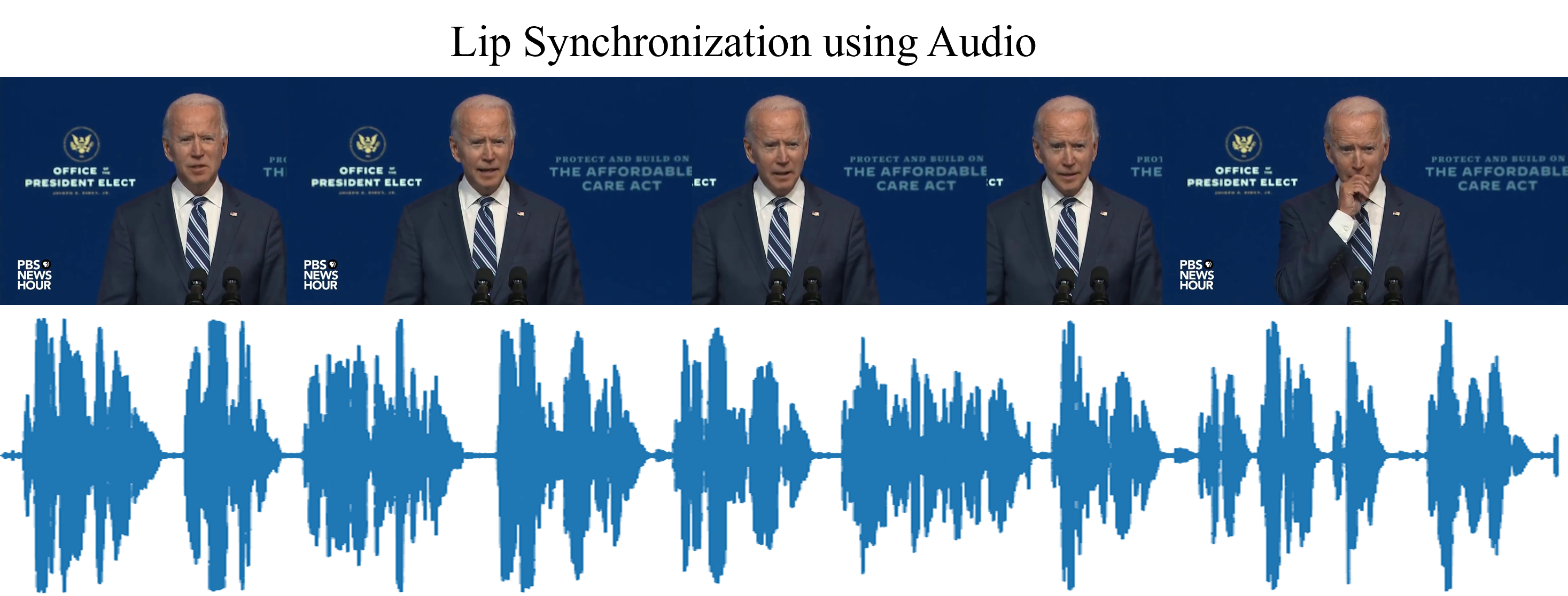}}} \vspace{-5pt}
    \caption{Examples of deepfake generation using different manipulation techniques: (a) Face swapping, where the source face is replaced with a target face in the FF++~\cite{rossler2019faceforensics++}/FSW dataset, and (b) lip synchronization, where facial lip movements are manipulated using different audio tracks in the PDD~\cite{sankaranarayanan2021presidential} dataset.}
    \label{fig:ana}
\end{figure}

\begin{figure}[!t]
    \centering
    {{\includegraphics[width=0.48\textwidth]{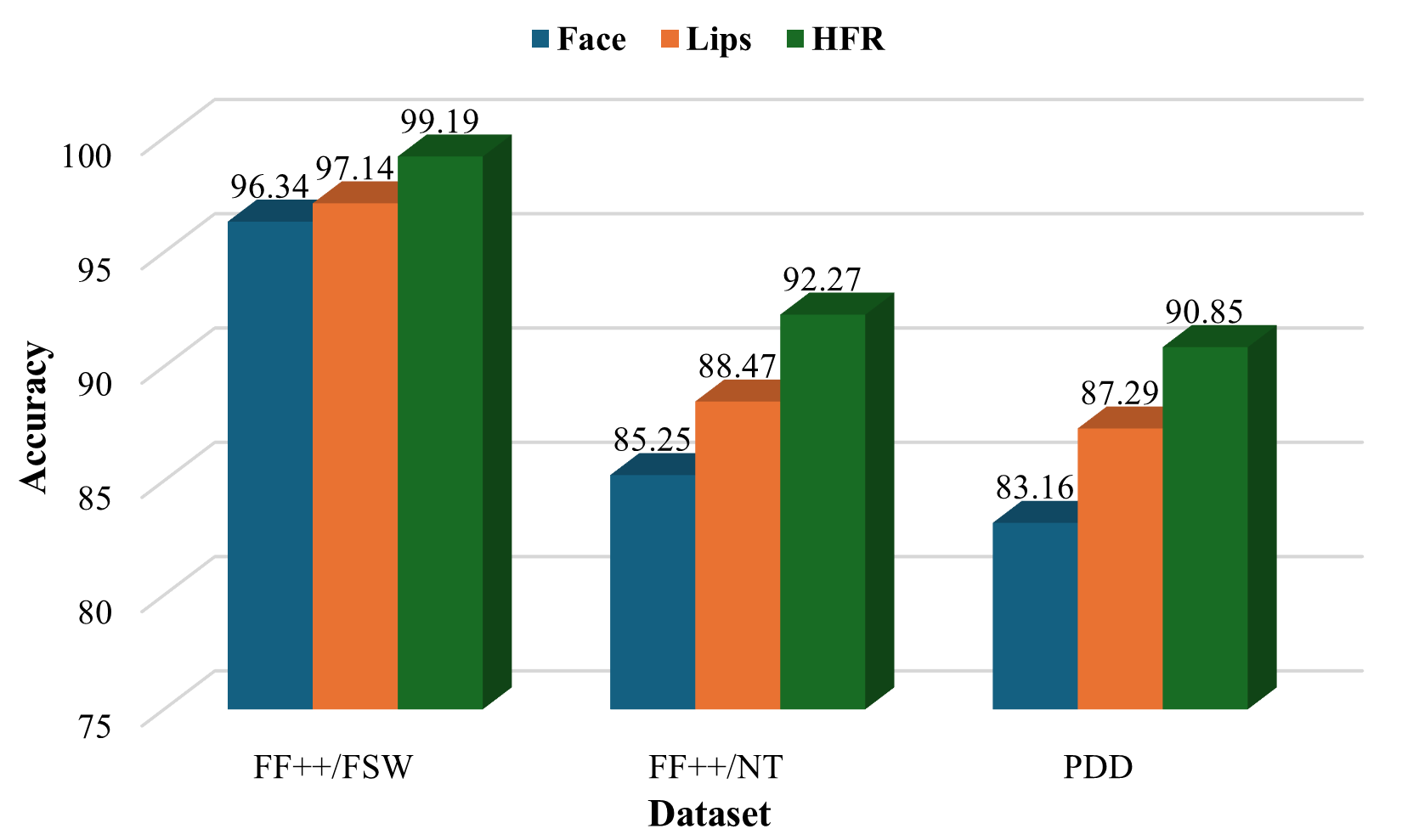} }}\vspace{-15pt}
    \caption{Performance analysis comparing feature representations: This study compares face-only and lips-only features with the proposed HFR approach. The results demonstrate that hierarchical feature representations provide superior effectiveness for deepfake detection across the FF++~\cite{rossler2019faceforensics++}/FSW, FF++~\cite{rossler2019faceforensics++}/NT, and PDD~\cite{sankaranarayanan2021presidential} datasets.}
    \vspace{-10pt}
    \label{fig:ana_3}
\end{figure}

\begin{figure*}[!t]
\vspace{-10pt}
    \centering
    {{\includegraphics[width=1\textwidth]{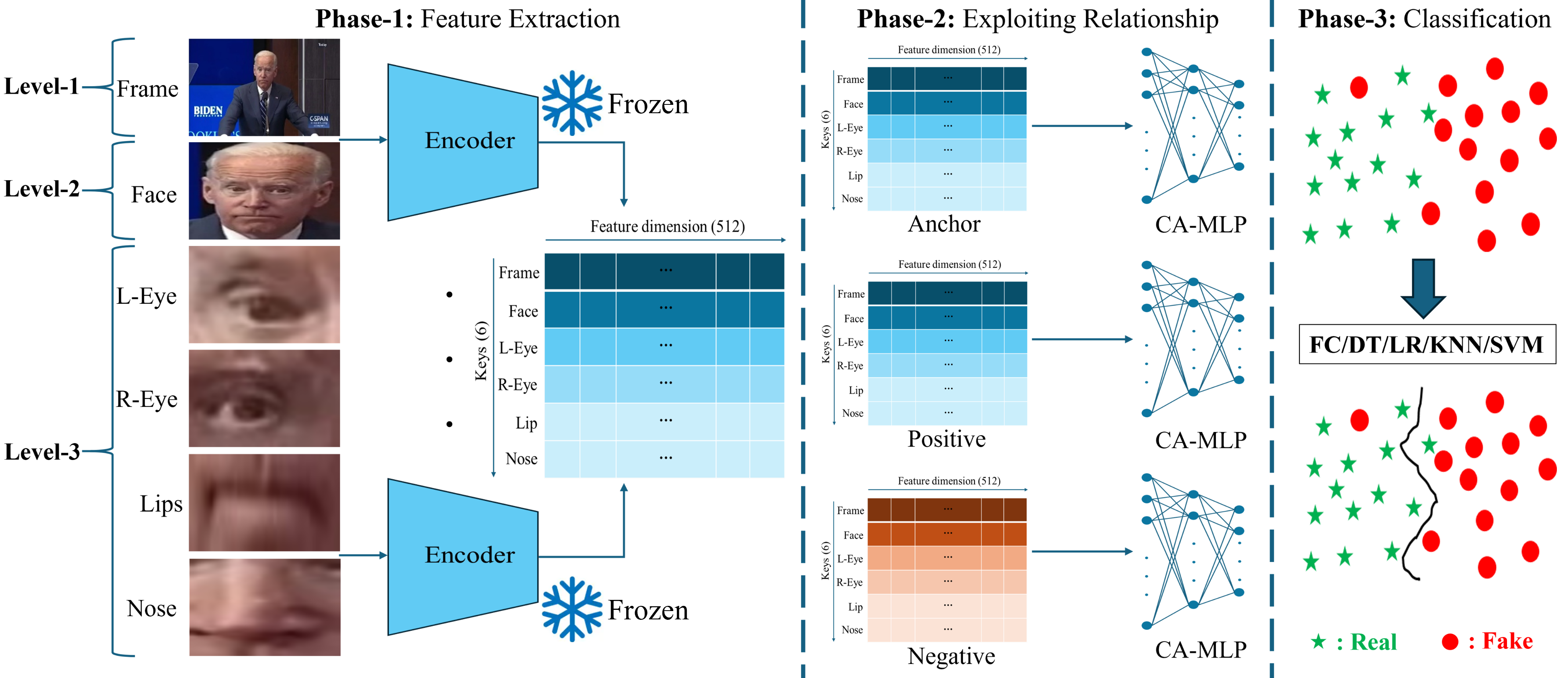}}}\vspace{-5pt}
    \caption{Architecture of the proposed method: The proposed framework is composed of three sequential phases. In Phase 1, an encoder network is employed to extract hierarchical feature representations from multiple facial regions, including the full frame, face, lips, left eye, right eye, and nose. In Phase 2, a deep triplet learning strategy combined with a CA-MLP mechanism is introduced to model and exploit interrelationships among regional features, yielding discriminative embeddings organized into two clusters. Finally, in Phase 3, the learned embeddings are passed to a classifier for final prediction.}\vspace{-10pt}
    \label{fig:arc}
\end{figure*}

\begin{figure}[!t]
    \centering
    {{\includegraphics[width=0.48\textwidth]{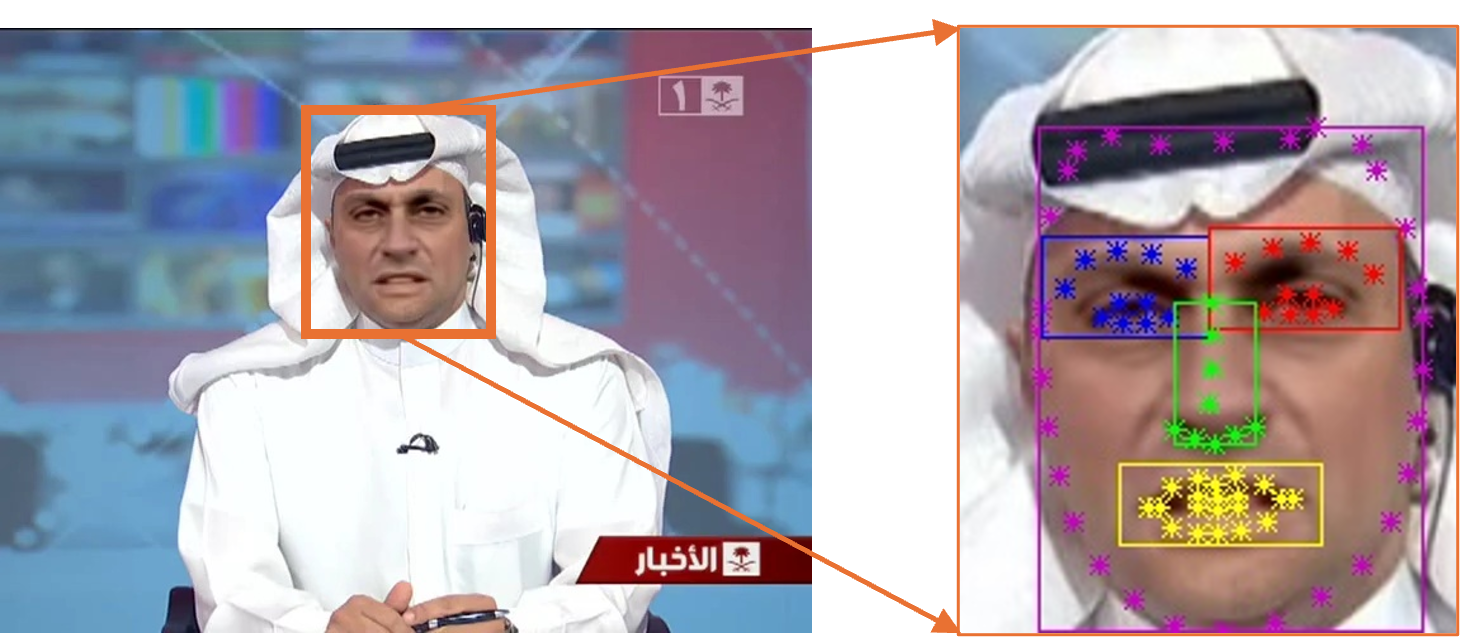} }} \vspace{-10pt}
    \caption{Key facial regions extracted using Dlib: An 81-point facial landmark detector was employed to accurately localize facial regions and capture detailed facial information. The colored rectangles indicate the selected facial regions used for extracting coarse-to-fine-grained hierarchical features.}
    \vspace{-15pt}
    \label{fig:keys}
\end{figure}

\section{Proposed Method}
\label{sec:method}
This section outlines the rationale behind the proposed method, focusing on its motivation, analysis, hierarchical feature extraction, and deep metric learning.

\subsection{Problem Statement}
Formally, let $I \in \mathbb{R}^{H \times W \times C}$ be an input image, where $H$, $W$, and $C$ indicate height, width, and channels, respectively. Most existing deepfake detection methods rely on extracting a medium-grained facial representation $F_{Face} = \Phi(I)$, where $\Phi(\cdot)$ denotes a feature extraction function that primarily focuses on the face region alone. However, if the deepfake is generated by selectively modifying a specific region $R_i$ (e.g., lips, eyes, or nose), then relying solely on $F_{Face}$ may fail to capture the fine-grained manipulations, leading to a degraded detection performance:
\begin{equation}
D(F_{Face}) = \sigma(W \cdot F_{Face} + b)
\end{equation}
where $D(\cdot)$ and  $\sigma(\cdot)$ indicate deepfake detection and activation function, and $W$ and $b$ are learnable parameters that depend on the inputs, hidden layers, and outputs.\\
To overcome these limitations, we propose a hybrid deepfake detection approach based on $HFR$, integrating multi-level feature extraction from different facial regions. The proposed approach leverages hierarchical relationships by incorporating coarse-to-fine information:
\begin{equation}
F_{HFR} = [\Phi(I_{L_i})]; i = 0...,n
\end{equation}
where $\Phi_{L_i}(\cdot)$ is the feature extraction function for the frame, face, and specific fine-grained regions (e.g., lips, left eye, right eye, nose). By leveraging a channel-attention mechanism and deep triplet learning, the proposed method captures inter-dependencies among them, allowing it to generalize effectively across various manipulations.\\
Thus, if an unseen sample $I^{'} \in \mathbb{R}^{H \times W \times C}$ is generated by selectively modifying a specific region $L_i$, the proposed $HFR$ ensures that the classifier retains sensitivity to coarse-to-fine-grained details, leading to robust detection:
\begin{equation}
D(F_{HFR}) = \sigma(W \cdot F_{HFR} + b)
\end{equation}

\subsection{Analysis}
As described in Section~\ref{sec:related}, most SoTA methods~\cite{matern2019exploiting, passos2024review, nataraj2019detecting}, ranging from traditional hand-crafted to hybrid techniques, utilize a medium-grain facial region for deepfake detection. However, such approaches often fail or perform poorly when the deepfake manipulations are localized to specific facial regions, such as the lips, eyes, or nose. We analyzed several SoTA methods trained on different manipulations, using the face alone for detection in the FF++~\cite{rossler2019faceforensics++}/FaceSwap (FF++~\cite{rossler2019faceforensics++}/FSW), FF++~\cite{rossler2019faceforensics++}/NeuralTextures (FF++~\cite{rossler2019faceforensics++}/NT), and the lip synchronization in the Presidential Deepfake Dataset (PDD~\cite{sankaranarayanan2021presidential}), as illustrated in Figure~\ref{fig:ana}. For FF++~\cite{rossler2019faceforensics++}/FSW, FF++~\cite{rossler2019faceforensics++}/NT, and PDD~\cite{sankaranarayanan2021presidential},  interestingly, focusing solely on the lip region provides better results than using the face alone. We reported the results using CLIP: ViT-B32~\cite{cherti2023reproducible}.\\
Furthermore, SoTA methods that rely solely on the face have limited capability to explore relationships among different facial regions. In contrast, the proposed $HFR$ utilizes each facial region separately, such as the left eye, right eye, lips, and nose, forcing it to learn the intricate relationships among them.
The proposed $HFR$ incorporates coarse(frame)-, medium(face)-, and fine(left eye, right eye, lips, and nose)-grained information from individual facial regions. By doing so, the model is guided to focus not only on the face but also on the broader context to capture both local and global discriminative features. This significantly improves overall performance as shown in Figure~\ref{fig:ana_3}.

\subsection{Architectural Overview}
Figure~\ref{fig:arc} illustrates the overall architecture of the proposed method, which is structured into three phases. The following subsections provide a detailed description of each phase.

\subsubsection{Hierarchical Feature Representation}
$HFR$ involves extracting discriminative features from various levels of information, ranging from coarse-to-fine-grained details. This includes analyzing the frame, face, left eye, right eye, lips, and nose separately to effectively capture diverse discriminative patterns for deepfake detection.\\
Let $I_{Frame}$ represent the input frame. First, we crop key facial regions, such as $I_{Face}$, $I_{Left-Eye}$, $I_{Right-Eye}$, $I_{Lips}$, and $I_{Nose}$  as shown in Figure~\ref{fig:keys} with different colors. We use Dlib with 81 facial landmarks to extract facial regions.
Let us assume that $F_{L_i}$ is a feature vector list of $I_{L_i}$ such as $F_{L}$ = [$F_{Frame}$, $F_{Face}$, $F_{Left-Eye}$, $F_{Right-Eye}$, $F_{Lips}$, $F_{Nose}$]. Next, we use a pre-trained encoder ($E$) model to extract feature, denoted as $F_{L_i}$, at level $i$ for $I_{L_i}$, defined as follow:
\begin{equation}
F_{L_i} = E(I_{L_i})
\end{equation}
We stack the extracted features and use them for further processing, defined as follows:
\begin{equation}
\begin{aligned}
[F_{L_1}, \dots, F_{L_{i-1}}] \text{ append } F_{L_{i}} = [F_{L_1}, \dots, F_{L_{i}}]
\end{aligned}
\end{equation}
which provides a six-by-$D$-dimensional feature map for three levels. $D$ is the dimension of each feature map. While $i=1$ and $i=2$ indicating level-1 and level-2, there are two feature maps ($F_{Frame}$ and $F_{Face}$) for $I_{Frame}$ and $I_{Face}$, respectively. While $i=3$ indicating level-3, there are four feature maps ($F_{Left-Eye}$, $F_{Right-Eye}$, $F_{Lips}$, and $F_{Nose}$) for $I_{Left-Eye}$, $I_{Right-Eye}$, $I_{Lips}$, and $I_{Nose}$, respectively.
\begin{figure}[t!]
    \centering
    {{\includegraphics[width=0.48\textwidth]{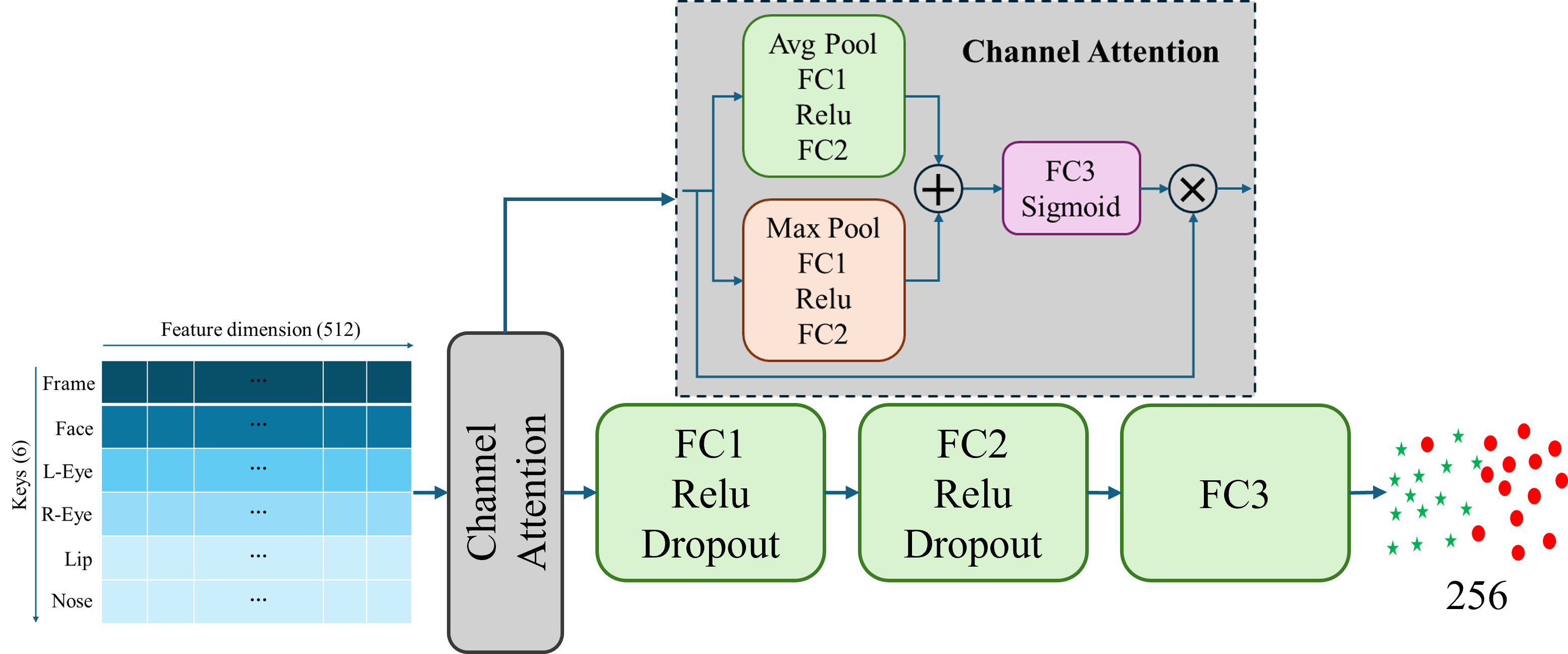}}}
    \caption{Block diagram of the proposed $CA$-MLP embedding network, which models relationships among multi-level hierarchical features extracted from the frame, face, left eye, right eye, lips, and nose regions.}
    \label{fig:ca_mlp}
    \vspace{-15pt}
\end{figure}

\begin{figure*}[t!]
    \centering
    {{\includegraphics[width=1\textwidth]{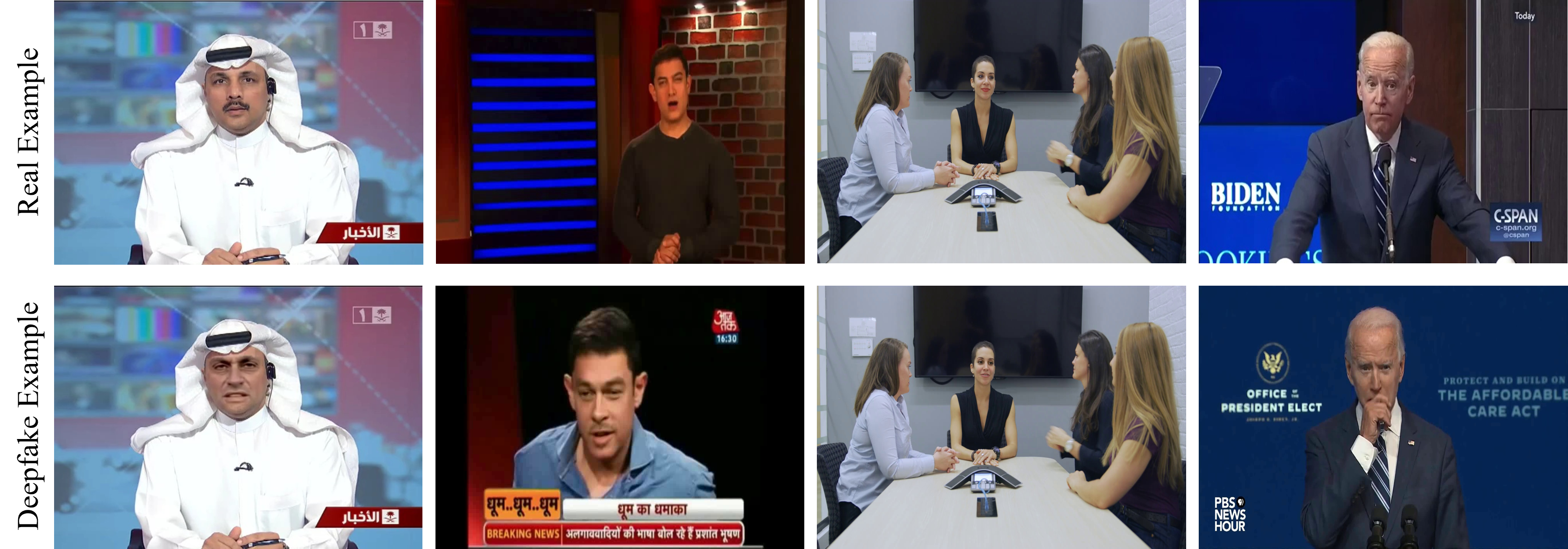}}}
    \caption{Illustrative examples of real and deepfake frames (from left to right) from the FF++~\cite{rossler2019faceforensics++}, CDF2~\cite{li2020celeb}, DFD~\cite{dolhansky2020deepfake}, and PDD~\cite{sankaranarayanan2021presidential} datasets. Deepfake faces in these datasets were generated using different synthesis models.}
    \vspace{-10pt}
    \label{fig:dataset}
\end{figure*}
\begin{table*}[!b]
\centering
\vspace{-8pt}
\caption{Detailed overview of the dataset: Including dataset name, number of subjects, target group characteristics, sample count, and manipulation types.}
\begin{tabular}{llllll}
\midrule \midrule
\textbf{Datasets} & \textbf{Subjects} & \textbf{Focus} & \textbf{\# of Samples (Real/Fake)} & \textbf{Manipulations} & \textbf{Train-Test} \\
\midrule
FF++~\cite{rossler2019faceforensics++} & 977 & Random & 1000/5000 & FaceSwap, FaceShift, NeuralTextures & 3600, 700 \\
CDF1~\cite{li2020celeb} & N/A & Celebrities & 408/795 & FaceSwap & 1027, 176 \\
CDF2~\cite{li2020celeb} & 59 & Celebrities & 590/5639 & FaceSwap & 5632, 597 \\
DFD~\cite{dolhansky2020deepfake} & 28 & Random & 363/3068 & Not disclosed, probably FaceSwap & 2744, 687 \\
DFDC~\cite{dolhansky2020deepfake} & 66 & Random & 1131/4119 & Not disclosed, probably FaceSwap & 3675, 1575 \\
DTIM~\cite{korshunov2018deepfakes} & 64 & Random & 320/320 & FaceSwap & 448, 192 \\
PDD~\cite{sankaranarayanan2021presidential} & 2 & US President & 16/16 & Lip Synchronization & 20, 12 \\
WLDR~\cite{agarwal2019protecting} & 6 & World Leaders & 595/87 & FaceSwap & 477, 205 \\
\midrule \midrule
\end{tabular}
\label{tab:dataset}
\end{table*}

\begin{table}[!b]
\centering
\vspace{-15pt}
\caption{Intra-dataset performance evaluations using spatial encoders. Results are presented for individual face regions and $HFR$.} 
\begin{adjustbox}{max width=0.5\textwidth}
\begin{tabular}{lcccccc}
\midrule \midrule
& \multicolumn{2}{c}{\textbf{ViT-B16~\cite{cherti2023reproducible}}} & \multicolumn{2}{c}{\textbf{ViT-B32~\cite{cherti2023reproducible}}} & \multicolumn{2}{c}{\textbf{ViT-L14~\cite{cherti2023reproducible}~\cite{cherti2023reproducible}}} \\
\midrule \midrule
\textbf{Frame-level} & \textbf{Face} & \textbf{HFR} & \textbf{Face} & \textbf{HFR} & \textbf{Face} & \textbf{HFR} \\
\midrule
FF++~\cite{rossler2019faceforensics++}/DF & 97.32 & 99.37 & 96.77 & 99.38 & 96.88 & 99.28 \\
FF++~\cite{rossler2019faceforensics++}/F2F & 92.89 & 96.17 & 93.07 & 96.24 & 95.38 & 97.23 \\
FF++~\cite{rossler2019faceforensics++}/FSH & 95.67 & 98.93 & 97.02 & 98.75 & 96.34 & 98.84 \\
FF++~\cite{rossler2019faceforensics++}/FSW & 97.39 & 99.23 & 96.34 & 99.19 & 97.72 & 99.70 \\
FF++~\cite{rossler2019faceforensics++}/NT & 86.33 & 91.99 & 85.25 & 92.27 & 83.13 & 90.44 \\
\midrule
CDF1~\cite{li2020celeb} & 90.03 & 93.46 & 84.49 & 90.99 & 97.64 & 98.85 \\
CDF2~\cite{li2020celeb} & 91.91 & 95.61 & 89.69 & 94.26 & 96.37 & 97.66 \\
DFD~\cite{dolhansky2020deepfake} & 93.86 & 96.76 & 92.49 & 96.42 & 94.94 & 97.68 \\
DFDC~\cite{dolhansky2020deepfake} & 99.68 & 99.77 & 99.39 & 98.41 & 99.73 & 99.75 \\
DTIM~\cite{korshunov2018deepfakes} & 100.00 & 100.00 & 100.00 & 100.00 & 100.00 & 100.00 \\
PDD~\cite{sankaranarayanan2021presidential} & 78.29 & 86.79 & 83.16 & 90.85 & 89.89 & 92.37 \\
WLDR~\cite{agarwal2019protecting} & 100.00 & 100.00 & 100.00 & 100.00 & 100.00 & 100.00 \\
\midrule \midrule
\textbf{Video-level} & \textbf{Face} & \textbf{HFR} & \textbf{Face} & \textbf{HFR} & \textbf{Face} & \textbf{HFR} \\
\midrule
FF++~\cite{rossler2019faceforensics++}/DF & 98.66 & 99.89 & 98.22 & 99.84 & 98.04 & 99.86 \\
FF++~\cite{rossler2019faceforensics++}/F2F & 94.80 & 97.56 & 95.94 & 97.95 & 97.21 & 98.43 \\
FF++~\cite{rossler2019faceforensics++}/FSH & 98.16 & 99.56 & 98.99 & 99.46 & 98.35 & 99.47 \\
FF++~\cite{rossler2019faceforensics++}/FSW & 98.76 & 99.91 & 98.42 & 99.94 & 98.88 & 99.99 \\
FF++~\cite{rossler2019faceforensics++}/NT & 90.86 & 95.43 & 88.82 & 95.40 & 89.12 & 94.34 \\
\midrule
CDF1~\cite{li2020celeb} & 92.59 & 95.79 & 87.74 & 94.55 & 98.44 & 99.96 \\
CDF2~\cite{li2020celeb} & 95.77 & 98.47 & 95.39 & 97.51 & 98.67 & 99.45 \\
DFD~\cite{dolhansky2020deepfake} & 97.77 & 99.28 & 96.77 & 98.99 & 98.86 & 99.57 \\
DFDC~\cite{dolhansky2020deepfake} & 99.93 & 99.96 & 99.90 & 99.45 & 99.97 & 99.92 \\
DTIM~\cite{korshunov2018deepfakes} & 100.00 & 100.00 & 100.00 & 100.00 & 100.00 & 100.00 \\
PDD~\cite{sankaranarayanan2021presidential} & 83.33 & 91.67 & 80.56 & 91.67 & 94.44 & 94.44 \\
WLDR~\cite{agarwal2019protecting} & 100.00 & 100.00 & 100.00 & 100.00 & 100.00 & 100.00 \\
\midrule \midrule
\end{tabular}
\end{adjustbox}
\label{tab:intra}
\end{table}

\begin{table}[!b]
\vspace{-10pt}
\centering
\caption{Inter-dataset performance evaluations using spatial encoders. The model is trained using the FF++~\cite{rossler2019faceforensics++} dataset and tested on other datasets.}
\begin{adjustbox}{max width=0.5\textwidth}
\begin{tabular}{lcccccc}
\midrule \midrule
 & \multicolumn{2}{c}{\textbf{ViT-B16~\cite{cherti2023reproducible}}} & \multicolumn{2}{c}{\textbf{ViT-B32~\cite{cherti2023reproducible}}} & \multicolumn{2}{c}{\textbf{ViT-L14~\cite{cherti2023reproducible}~\cite{cherti2023reproducible}}} \\
 \midrule
\textbf{Frame-level} & \textbf{Face} & \textbf{HFR} & \textbf{Face} & \textbf{HFR} & \textbf{Face} & \textbf{HFR} \\
\midrule \midrule
CDF1~\cite{li2020celeb} & 64.53 & 68.68 & 63.41 & 62.90 & 73.02 & 76.76 \\
CDF2~\cite{li2020celeb} & 67.04 & 71.75 & 68.70 & 72.25 & 74.20 & 78.63 \\
DFD~\cite{dolhansky2020deepfake} & 83.18 & 85.88 & 81.30 & 88.67 & 83.32 & 86.73 \\
DFDC~\cite{dolhansky2020deepfake} & 72.19 & 73.45 & 67.47 & 67.69 & 71.58 & 76.98 \\
DTIM~\cite{korshunov2018deepfakes} & 85.70 & 91.81 & 85.17 & 92.94 & 86.63 & 93.54 \\
PDD~\cite{sankaranarayanan2021presidential} & 64.69 & 81.30 & 63.62 & 68.70 & 60.32 & 69.00 \\
WLDR~\cite{agarwal2019protecting} & 77.49 & 95.70 & 95.62 & 98.68 & 91.91 & 98.78 \\
\midrule \midrule
\textbf{Video-level} & \textbf{Face} & \textbf{HFR} & \textbf{Face} & \textbf{HFR} & \textbf{Face} & \textbf{HFR} \\
\midrule \midrule
CDF1~\cite{li2020celeb} & 70.03 & 76.64 & 70.27 & 78.55 & 75.52 & 79.80 \\
CDF2~\cite{li2020celeb}  & 72.06 & 76.71 & 74.95 & 77.23 & 81.59 & 85.34 \\
DFD~\cite{dolhansky2020deepfake}  & 88.35 & 88.65 & 87.07 & 92.78 & 88.29 & 89.41 \\
DFDC~\cite{dolhansky2020deepfake}  & 76.13 & 78.73 & 71.12 & 74.31 & 80.99 & 84.07 \\
DTIM~\cite{korshunov2018deepfakes}  & 88.47 & 94.14 & 88.57 & 96.47 & 91.88 & 95.62 \\
PDD~\cite{sankaranarayanan2021presidential}  & 69.44 & 80.56 & 58.33 & 75.00 & 66.67 & 80.76 \\
WLDR~\cite{agarwal2019protecting}  & 79.33 & 97.12 & 99.04 & 100.00 & 98.56 & 100.00 \\
\midrule \midrule
\end{tabular}
\end{adjustbox}
\label{tab:inter}
\end{table}

\begin{table*}[]
\centering
\vspace{-10pt}
\caption{Performance comparisons of the proposed method with the SoTA methods. All models are trained on FF++~\cite{rossler2019faceforensics++} and tested on others. We compared the proposed method for spatial encoders with SoTA mehtods.}
\begin{adjustbox}{max width=1.5\textwidth}
\begin{tabular}{ccccccccc}
\midrule \midrule
\textbf{Methods} & FF++~\cite{rossler2019faceforensics++} & CDF1~\cite{li2020celeb} & CDF2~\cite{li2020celeb} & DFD~\cite{dolhansky2020deepfake} & DFDC~\cite{dolhansky2020deepfake} & DTIM~\cite{korshunov2018deepfakes} & PDD~\cite{sankaranarayanan2021presidential} & WLDR~\cite{agarwal2019protecting} \\
\midrule
PBD~\cite{schwarcz2021finding}, 2021  & 93.10 & - & 63.30 & - & 62.70 & - & - & - \\
{CORE~\cite{ni2022core}, 2022} & 96.38 & 77.98 & 74.28 & 80.18 & 70.49 & 68.39 & 58.33 & 86.96 \\
{Recce~\cite{cao2022end}, 2022} & 96.21 & 76.77 & 73.19 & 81.19 & 71.33 & 85.83 & \underline{75.00} & 81.82 \\
SSL~\cite{chen2022self}, 2022 & 98.40 & - & 79.70 & 77.20 & - & - & - & - \\
EfficientB4\cite{yan2023deepfakebench}, 2023 & 95.67 & 79.09 & 74.87 & 81.48 & 69.55 & 64.78 & 58.33 & 54.54 \\
Xception\cite{yan2023deepfakebench}, 2023 & 96.37 & 77.94 & 73.65 & 81.63 & 70.77 & 57.08 & 66.67 & 83.94 \\
SFGD~\cite{wang2023dynamic}, 2023 & 95.98 & - & 75.83 & 88 & 73.64 & - & - & - \\
F3Net\cite{yan2023deepfakebench}, 2023 & 96.35 & 77.69 & 73.52 & 79.75 & 70.21 & 63.54 & 66.67 & \underline{87.88} \\
{UCF~\cite{yan2023ucf}, 2023} & 97.05 & 77.93 & 75.27 & 80.74 & 71.91 & 55.58 & 66.67 & 86.97 \\
{PFG~\cite{lin2024preserving}, 2024} & 98.28 & - & 74.42 & 84.82 & 61.47 & - & - & - \\
{GM-DF~\cite{lai2024gm}, 2024} & 96.62 & - & 83.16 & - & 77.23 & - & - & - \\
LSDA~\cite{lai2024gm}, 2024 & - & \textbf{86.70} & 83.00 & 88.00 & 73.60 & - & - & - \\
{DSM~\cite{zhang2025dsm}, 2025} & 99.0 & - & \textbf{91.27} & \underline{88.14} & \underline{80.81} & - & - & - \\
{EFFT~\cite{yan2024orthogonal}, 2025} & 97.20 & - & 89.45 & 86.98 & 77.65 & \underline{90.72} & 67.31 & 95.23 \\
{D3~\cite{zheng2025d3}, 2025} & 96.24 & - & 74.55 & 79.12 & 68.29 & 74.13 & 58.91 & 88.26 \\
\midrule
{\textbf{HFR}} & \underline{98.42} & \underline{79.80} & \underline{85.34} & \textbf{89.41} & \textbf{84.07} & \textbf{95.62} & \textbf{80.76} & \textbf{100} \\
\midrule \midrule
\end{tabular}
\end{adjustbox}
\label{tab:per_comparison}
\end{table*}

\begin{table}[]
\centering
\caption{Performance comparisons of inter-manipulation detection in the FF++~\cite{rossler2019faceforensics++} dataset using a spatial encoder model (CLIP: ViT-L14~\cite{cherti2023reproducible}), which includes four types of manipulations.}
\begin{adjustbox}{max width=0.47\textwidth}
\begin{tabular}{llllll}
\midrule \midrule
 &  & DF & F2F & FSW & NT \\
 \midrule
\multirow{5}{*}{DF} 
& MAT~\cite{zhao2021multi}, 2021 & 99.92 & 75.23 & 40.61 & 71.08 \\
 & GFF~\cite{luo2021generalizing}, 2021 & 99.87 & 76.89 & 47.21 & 72.88 \\
 & DCL~\cite{sun2022dual}, 2022 & \underline{99.98} & 77.13 & 61.01 & 75.01 \\
 & DSM~\cite{zhang2025dsm}, 2025 & \textbf{99.99} & \underline{79.59} & \underline{62.62} & \underline{76.8} \\
 \cline{2-6}
 & \textbf{Proposed HFR} & 99.86 & \textbf{81.62} & \textbf{93.13} & \textbf{79.58} \\
 \midrule  \midrule
\multirow{5}{*}{F2F} 
& MAT~\cite{zhao2021multi}, 2021 & 86.15 & 99.13 & 60.14 & 64.59 \\
 & GFF~\cite{luo2021generalizing}, 2021 & 89.23 & 99.1 & 61.3 & 64.77 \\
 & DCL~\cite{sun2022dual}, 2022 & \underline{91.91} & \underline{99.21} & 59.58 & 66.67 \\
 & DSM~\cite{zhang2025dsm}, 2025 & 90.01 & \textbf{99.3} & \underline{63.6} & \underline{67.76} \\
 \cline{2-6}
 & \textbf{Proposed HFR} & \textbf{95.34} & 98.43 & \textbf{93.17} & \textbf{77.86} \\
 \midrule  \midrule
\multirow{5}{*}{FSW} 
& MAT~\cite{zhao2021multi}, 2021 & 64.13 & 66.39 & 99.67 & 50.1 \\
 & GFF~\cite{luo2021generalizing}, 2021 & 70.21 & 68.72 & 99.85 & 49.91 \\
 & DCL~\cite{sun2022dual}, 2022 & 74.8 & 69.75 & \underline{99.9} & 52.6 \\
 & DSM~\cite{zhang2025dsm}, 2025 & 85.6 & \underline{70.19} & 99.76 & \underline{53.65} \\
 \cline{2-6}
 & \textbf{Proposed HFR} & \textbf{92.54} & \textbf{87.46} & \textbf{99.99} & \textbf{64.56} \\
 \midrule  \midrule
\multirow{5}{*}{NT} 
 & MAT~\cite{zhao2021multi}, 2021 & 87.23 & 48.22 & 75.33 & 98.66 \\
 & GFF~\cite{luo2021generalizing}, 2021 & 88.49 & 49.81 & 74.31 & \underline{98.77} \\
 & DCL~\cite{sun2022dual}, 2022 & 91.23 & 52.13 & \textbf{79.31} & \textbf{98.97} \\
 & DSM~\cite{zhang2025dsm}, 2025 & \underline{92.74} & \underline{60.88} & \underline{79.29} & 98.52 \\
 \cline{2-6}
 & \textbf{Proposed HFR} & \textbf{96.31} & \textbf{77.24} & 66.67 & 94.34 \\
 \midrule \midrule
\end{tabular}
\end{adjustbox}
\label{tab:inter_comparison}
\vspace{-25pt}
\end{table}

\subsubsection{Exploiting Relationships with Triplet Learning}
Triplet learning~\cite{hoffer2015deep} is commonly used in computer vision to exploit relationships among data samples by capturing similarities within classes and dissimilarities between them. In deepfake detection, it enhances the model's ability to distinguish real from fake content by improving the representation of both classes, thus boosting detection performance.\\
In the proposed method, we leverage three levels of representative features, utilizing triplet learning to find relationships among them, as illustrated in Phase 2 of Figure~\ref{fig:arc}. To achieve this, we design a channel-attention-based multi-layer perceptron ($CA-MLP$) model, referred to as the triplet ($T$) network, which incorporates a channel-attention block along with three additional blocks, as illustrated in Figure~\ref{fig:ca_mlp}. The channel attention mechanism is employed to determine the most important feature levels for deepfake detection by performing both global average pooling and global max pooling on the features. Global average pooling captures the global importance of each feature, while max pooling highlights the most salient features, such as the frame, face, eyes, lips, and nose. The pooled features are passed through three fully connected layers to generate channel-wise attention weights, enabling the model to focus on the most informative features. After channel attention, the output passes through two fully connected layers to reduce the feature space to learn discriminative properties.\\
To train the triplet model, we randomly select an anchor ($A$) sample from either the real or fake class. A positive ($P$) sample is chosen from the same class, while a negative ($N$) sample is from the other class.
These samples are processed through a triplet ($T$) model to generate embeddings, $F_A$, $F_P$, and $F_N$, mathematically represented as:
\vspace{-3pt}
\begin{equation}
F_A, F_P, F_{N} = T(A, P, N)
\end{equation}
Then pairwise euclidean distance, $D_{AP}$ between $F_A$ and $F_P$, and distance $D_{AN}$ between $F_A$ and $F_N$ are computed. We use margin ranking loss ($\lambda$) to learn the discriminative properties of three levels of features, defined as follows:
\begin{equation}
\lambda = max(0, (D_{AP}-D_{AN})+margin)
\end{equation}
We examined margin values, such as 0, 0.1, 0.2, and 0.3, and set the margin to 0, which provides the best performance.\\
After training the triplet network, we integrate a classifier for final prediction, as shown in Phase-3 of Figure~\ref{fig:arc}. The $T$ model generates 256-dimensional discriminative features. We tested fully connected ($FC$), decision tree ($DT$), logistic regression ($LR$), k-nearest neighbor ($KNN$), support vector machine ($SVM$), gradient boosting ($GB$), and random forest ($RF$), and found that $FC$ outperformed the others.

\begin{table}[!t]
\centering
\caption{Intra-dataset performance evaluations for additional experiments using MViT~\cite{fan2021multiscale}, TVC~\cite{bertasius2021space}, and ViViT~\cite{arnab2021vivit}.}
\begin{adjustbox}{max width=0.5\textwidth}
\begin{tabular}{lcccccc}
\midrule \midrule
 & \multicolumn{2}{c}{\textbf{MViT~\cite{fan2021multiscale}}} & \multicolumn{2}{c}{\textbf{TVC~\cite{bertasius2021space}}} & \multicolumn{2}{c}{\textbf{ViViT~\cite{arnab2021vivit}}} \\
 \midrule \midrule
\textbf{Frame-level} & \textbf{Face} & \textbf{HFR} & \textbf{Face} & \textbf{HFR} & \textbf{Face} & \textbf{HFR} \\
\midrule
FF++~\cite{rossler2019faceforensics++}/DF & 91.92 & 99.32 & 97.77 & 99.37 & 97.49 & 99.28 \\
FF++~\cite{rossler2019faceforensics++}/F2F & 76.95 & 94.09 & 94.1 & 97.35 & 90.55 & 97.79 \\
FF++~\cite{rossler2019faceforensics++}/FSH & 80.4 & 97.08 & 95.59 & 98.62 & 96.59 & 99.1 \\
FF++~\cite{rossler2019faceforensics++}/FSW & 81.58 & 98.97 & 95.68 & 99.77 & 95.54 & 99.86 \\
FF++~\cite{rossler2019faceforensics++}/NT & 79.35 & 89.85 & 85.68 & 95.16 & 85.8 & 93.63 \\
\midrule
CDF1~\cite{li2020celeb} & 62.97 & 75.3 & 83.85 & 92.72 & 80.82 & 91.09 \\
CDF2~\cite{li2020celeb} & 69.13 & 86.25 & 90.41 & 97.23 & 87.41 & 96.76 \\
DFD~\cite{dolhansky2020deepfake} & 75.01 & 91.71 & 92.07 & 96.24 & 91.1 & 96.1 \\
DFDC~\cite{dolhansky2020deepfake} & 75.49 & 87.34 & 96.55 & 99.19 & 95.85 & 98.81 \\
DTIM~\cite{korshunov2018deepfakes} & 79.09 & 99.57 & 99.57 & 100 & 99.62 & 99.96 \\
PDD~\cite{sankaranarayanan2021presidential} & 74.68 & 89.34 & 81.36 & 82.89 & 67 & 91.26 \\
WLDR~\cite{agarwal2019protecting} & 96.01 & 99.99 & 99.5 & 100 & 99.87 & 100 \\
\midrule \midrule
\textbf{Video-level} & \textbf{Face} & \textbf{HFR} & \textbf{Face} & \textbf{HFR} & \textbf{Face} & \textbf{HFR} \\
\midrule
FF++~\cite{rossler2019faceforensics++}/DF & 98.25 & 99.96 & 98.88 & 99.87 & 98.55 & 99.7 \\
FF++~\cite{rossler2019faceforensics++}/F2F & 88.02 & 98.85 & 96.04 & 98.54 & 93.74 & 98.94 \\
FF++~\cite{rossler2019faceforensics++}/FSH & 91.24 & 99.37 & 96.98 & 98.92 & 97.87 & 99.38 \\
FF++~\cite{rossler2019faceforensics++}/FSW & 95 & 99.97 & 96.83 & 99.92 & 96.69 & 99.99 \\
FF++~\cite{rossler2019faceforensics++}/NT & 90.6 & 96.74 & 87.34 & 96.87 & 88.57 & 96.21 \\
\midrule
CDF1~\cite{li2020celeb} & 70.63 & 87.1 & 87.54 & 95.55 & 84.21 & 94.83 \\
CDF2~\cite{li2020celeb} & 77.49 & 96.27 & 94.06 & 98.78 & 92.82 & 98.9 \\
DFD~\cite{dolhansky2020deepfake} & 86.76 & 98.03 & 96.3 & 98.36 & 95 & 98.56 \\
DFDC~\cite{dolhansky2020deepfake} & 86.65 & 95.49 & 98.4 & 99.62 & 98.4 & 99.58 \\
DTIM~\cite{korshunov2018deepfakes} & 89.96 & 100 & 99.86 & 100 & 99.92 & 100 \\
PDD~\cite{sankaranarayanan2021presidential} & 77.78 & 100 & 86.11 & 86.11 & 69.44 & 91.67 \\
WLDR~\cite{agarwal2019protecting} & 100 & 100 & 99.52 & 100 & 100 & 100 \\
\midrule \midrule
\end{tabular}
\end{adjustbox}
\label{tab:ad_intra}
\vspace{-15pt}
\end{table}

\begin{table}[!t]
\centering
\caption{Inter-dataset performance evaluations. The model is trained using the FF++~\cite{rossler2019faceforensics++} dataset and tested on others}
\begin{adjustbox}{max width=0.5\textwidth}
\begin{tabular}{lcccccc}
\midrule \midrule
 & \multicolumn{2}{c}{\textbf{MViT~\cite{fan2021multiscale}}} & \multicolumn{2}{c}{\textbf{TVC~\cite{bertasius2021space}}} & \multicolumn{2}{c}{\textbf{ViViT~\cite{arnab2021vivit}}} \\
 \midrule \midrule
\textbf{Frame-level} & \textbf{Face} & \textbf{HFR} & \textbf{Face} & \textbf{HFR} & \textbf{Face} & \textbf{HFR} \\
\midrule 
CDF1~\cite{li2020celeb} & 60.77 & 61.92 & 68.37 & 70.80 & 66.46 & 74.13 \\
CDF2~\cite{li2020celeb} & 65.48 & 69.77 & 77.29 & 78.28 & 73.30 & 75.71 \\
DFD~\cite{dolhansky2020deepfake} & 69.08 & 84.45 & 82.37 & 88.84 & 82.65 & 88.23 \\
DFDC~\cite{dolhansky2020deepfake} & 63.12 & 65.23 & 63.90 & 65.93 & 64.94 & 60.39 \\
DTIM~\cite{korshunov2018deepfakes} & 68.15 & 89.38 & 81.64 & 95.52 & 81.46 & 96.14 \\
PDD~\cite{sankaranarayanan2021presidential} & 68.41 & 66.17 & 67.69 & 70.84 & 63.74 & 81.54 \\
WLDR~\cite{agarwal2019protecting} & 75.65 & 94.49 & 94.19 & 98.82 & 96.05 & 99.52 \\
\midrule \midrule
\textbf{Video-level} & \textbf{Face} & \textbf{HFR} & \textbf{Face} & \textbf{HFR} & \textbf{Face} & \textbf{HFR} \\
\midrule
CDF1~\cite{li2020celeb} & 67.83 & 69.87 & 69.83 & 73.48 & 68.31 & 76.36 \\
CDF2~\cite{li2020celeb} & 75.87 & 82.04 & 80.61 & 81.47 & 77.08 & 79.82 \\
DFD~\cite{dolhansky2020deepfake} & 76.18 & 90.67 & 82.92 & 89.28 & 84.16 & 89.26 \\
DFDC~\cite{dolhansky2020deepfake} & 69.56 & 71.94 & 65.72 & 67.22 & 61.56 & 64.94 \\
DTIM~\cite{korshunov2018deepfakes} & 74.65 & 97.36 & 81.51 & 95.97 & 82.05 & 97.83 \\
PDD~\cite{sankaranarayanan2021presidential} & 80.56 & 88.89 & 66.67 & 75 & 66.67 & 80.56 \\
WLDR~\cite{agarwal2019protecting} & 96.15 & 100 & 96.63 & 100 & 98.56 & 100 \\
\midrule \midrule
\end{tabular}
\end{adjustbox}
\label{tab:ad_inter}
\end{table}

\begin{table}[!t]
\centering
\caption{Inter-manipulations detection in FF++~\cite{rossler2019faceforensics++} datasets for additional experiments. We trained the model on one and tested it on the rest.}
\renewcommand{\arraystretch}{1} 
\begin{adjustbox}{max width=0.5\textwidth}
\begin{tabular}{l@{\hskip 3pt}l@{\hskip 3pt}lllll}
 \midrule \midrule
 \textbf{Frame-level} & & \textbf{DF} & \textbf{F2F} & \textbf{FSH} & \textbf{FSW} & \textbf{NT} \\
\midrule
\multirow{2}{*}{DF} & Face\hspace{5mm} & 97.77 & 69.37 & 65.75 & 49.01 & 70.14 \\
 & \textbf{HFR}\hspace{5mm} & \textbf{99.37} & \textbf{70.25} & \textbf{78.02} & \textbf{52.31} & \textbf{81.31} \\
 \midrule
\multirow{2}{*}{F2F} & Face~ & 79.82 & 94.10 & 51.79 & 60.42 & 63.54 \\
 & \textbf{HFR}~ & \textbf{80.79} & \textbf{97.35} & \textbf{52.65} & \textbf{52.54} & \textbf{74.40} \\
 \midrule
\multirow{2}{*}{FSH} & Face & 66.98 & 51.15 & 95.59 & 54.86 & 61.50 \\
 & \textbf{HFR} & \textbf{77.06} & \textbf{52.32} & \textbf{98.62} & \textbf{60.54} & \textbf{65.34} \\
 \midrule
\multirow{2}{*}{FSW} & Face & 59.56 & 56.80 & 56.68 & 95.68 & 39.85 \\
 & \textbf{HFR} & \textbf{63.27} & \textbf{56.11} & \textbf{62.11} & \textbf{99.77} & \textbf{54.71} \\
 \midrule
\multirow{2}{*}{NT} & Face & 76.62 & 61.53 & 64.97 & 45.08 & 85.68 \\
 & \textbf{HFR} & \textbf{94.85} & \textbf{76.00} & \textbf{78.20} & \textbf{51.40} & \textbf{95.16} \\
\midrule \midrule
 \textbf{Video-level} & & \textbf{DF} & \textbf{F2F} & \textbf{FSH} & \textbf{FSW} & \textbf{NT} \\
\midrule
\multirow{2}{*}{DF} & Face & 98.88 & 71.35 & 69.43 & 50.21 & 73.37 \\
 & \textbf{HFR} & \textbf{99.87} & \textbf{72.08} & \textbf{80.40} & \textbf{62.18} & \textbf{84.27} \\
 \midrule
\multirow{2}{*}{F2F} & Face & 83.36 & 96.04 & 52.48 & 62.03 & 65.53 \\
 & \textbf{HFR} & \textbf{83.04} & \textbf{98.54} & \textbf{59.27} & \textbf{58.87} & \textbf{77.64} \\
 \midrule
\multirow{2}{*}{FSH} & Face & 68.14 & 50.55 & 96.98 & 54.08 & 62.14 \\
 & \textbf{HFR} & \textbf{79.15} & \textbf{50.59} & \textbf{98.92} & \textbf{62.35} & \textbf{66.59} \\
 \midrule
\multirow{2}{*}{FSW} & Face & 61.53 & 57.41 & 58.30 & 96.83 & 37.89 \\
 & \textbf{HFR} & \textbf{30.41} & \textbf{56.66} & \textbf{63.51} & \textbf{99.92} & \textbf{44.67} \\
 \midrule
\multirow{2}{*}{NT} & Face & 78.56 & 61.12 & 65.26 & 44.60 & 87.34 \\
 & \textbf{HFR} & \textbf{96.75} & \textbf{79.10} & \textbf{80.17} & \textbf{51.51} & \textbf{96.87} \\
\midrule \midrule
\end{tabular}
\end{adjustbox}
\label{tab:ad_inter_man}
\end{table}

\begin{table}[!t]
\centering
\caption{Complexity Analysis.}
\begin{adjustbox}{max width=0.48\textwidth}
\begin{tabular}{lccc}
 \midrule \midrule
\textbf{Methods} & \textbf{Parameters (M)} & \textbf{FLOPs (G)} & \textbf{Time (S)} \\
\midrule
ViT-B16~\cite{cherti2023reproducible} + CA-MLP & 88.29 & 19.36 & 0.129 \\
ViT-B32~\cite{cherti2023reproducible} + CA-MLP & 89.95 & 6.86 & 0.143 \\
ViT-L14~\cite{cherti2023reproducible} + CA-MLP & 306.46 & 81.12 & 0.162 \\
\midrule
MViT~\cite{fan2021multiscale} + CA-MLP & 38.79 & 58.96 & 0.168 \\
TVC~\cite{bertasius2021space} + CA-MLP & 123.75 & 762.14 & 0.273 \\
ViViT~\cite{arnab2021vivit} + CA-MLP & 91.14 & 272.95 & 0.201 \\
 \midrule \midrule
\end{tabular}
\end{adjustbox}
\label{tab:complex}
\vspace{-15pt}
\end{table}

\section{Results and Discussion}
\label{sec:results}
This section covers the implementation details, dataset, performance evaluations, comparisons, and ablation studies.
\subsection{Datasets}
We conducted experiments on the proposed methods using several benchmark datasets: FF++~\cite{rossler2019faceforensics++}, Celeb-DF-V1 (CDF1~\cite{li2020celeb}), Celeb-DF-V2 (CDF2~\cite{li2020celeb}), Deepfake Detection (DFD~\cite{dolhansky2020deepfake}), Deepfake Detection Challenge (DFDC~\cite{dolhansky2020deepfake}), DeepfakeTIMIT (DTIM~\cite{korshunov2018deepfakes}), PDD~\cite{sankaranarayanan2021presidential}, and World Leaders (WLDR~\cite{agarwal2019protecting}). We followed the same train-test split scenario as reported in~\cite{yan2023deepfakebench} for all the datasets.\\
Table~\ref{tab:dataset} provides the dataset names, the number of subjects involved in creating each dataset, the primary focus of each dataset, the total number of real and deepfake videos, and the types of manipulations performed. For better comparison, we present both real and deepfake examples as shown in Figure~\ref{fig:dataset} for FF++~\cite{rossler2019faceforensics++}, CDF2~\cite{li2020celeb}, DFD~\cite{dolhansky2020deepfake}, and PDD~\cite{sankaranarayanan2021presidential} datasets, respectively. Regarding preprocessing consistency, we followed the same pipeline across all datasets: facial regions were extracted using Dlib~\cite{king2009dlib} with 81-point landmarks, resized to $224\times224$, and preprocessed using the CLIP~\cite{cherti2023reproducible} preprocessor before feature extraction. This uniform pipeline, consistent with DeepfakeBench, was applied to all datasets regardless of their source, ensuring fair cross-dataset evaluation. To mitigate data imbalance during triplet training, anchor, positive, and negative samples are systematically sampled from both real and fake classes. This strategy ensures balanced pairwise supervision at the triplet level, independent of overall dataset-level class imbalance.

\subsection{Encoder Models}
We use three pre-trained models to extract features: CLIP ViT-B16~\cite{cherti2023reproducible}, ViT-B32~\cite{cherti2023reproducible}, and ViT-L14~\cite{cherti2023reproducible}. We select CLIP models based on their effectiveness, generalization, and available source code, as the lack of source code can hinder result reproduction. CLIP models, known for their effective feature representation and generalizability in image processing and computer vision, are promising candidates for deepfake detection.\\
We conducted the entire experiment on a Linux 22.04 system equipped with two GPUs, each with 46GB of memory. We cropped each facial region using Dlib and resized it to $224 \times 224$ to extract features using pretrained encoder models. Before feature extraction, we applied preprocessing using the CLIP preprocessor to ensure compatibility with the encoder models. We trained a triplet model for 20 epochs with a batch size of 100 and a learning rate of 0.0001 to enhance feature representation. The Adam optimizer was employed to optimize the triplet model.

\subsection{Evaluation Metric}
\label{sec:metric}
We evaluate the performance of the proposed method using the area under the receiver operating characteristic curve (AUC), which is widely used for binary classification tasks and is robust to class imbalance. AUC measures the ability of a model to distinguish between real and fake samples across all decision thresholds.\\
First, we compute the frame-level AUC (F-AUC). Let $p_i$ denote the predicted probability for frame $i$, and $y_i \in \{0,1\}$ denote the corresponding ground-truth label. The F-AUC is calculated using the ROC curve obtained from the frame-level predictions.\\
Next, we compute the video-level AUC (V-AUC). For each video $j$, we aggregate the predictions of all frames by averaging their probabilities:
\begin{equation}
P_j = \frac{1}{T_j} \sum_{i=1}^{T_j} p_{j,i}
\end{equation}
where $T_j$ denotes the number of frames in video $j$. The resulting video-level score $P_j$ is then used to compute the ROC curve and the corresponding V-AUC.
\begin{figure*}[!t]
    \centering
    \vspace{-15pt}
    \subfloat[\label{fig:frame_idx}]{\includegraphics[width=0.49\textwidth]{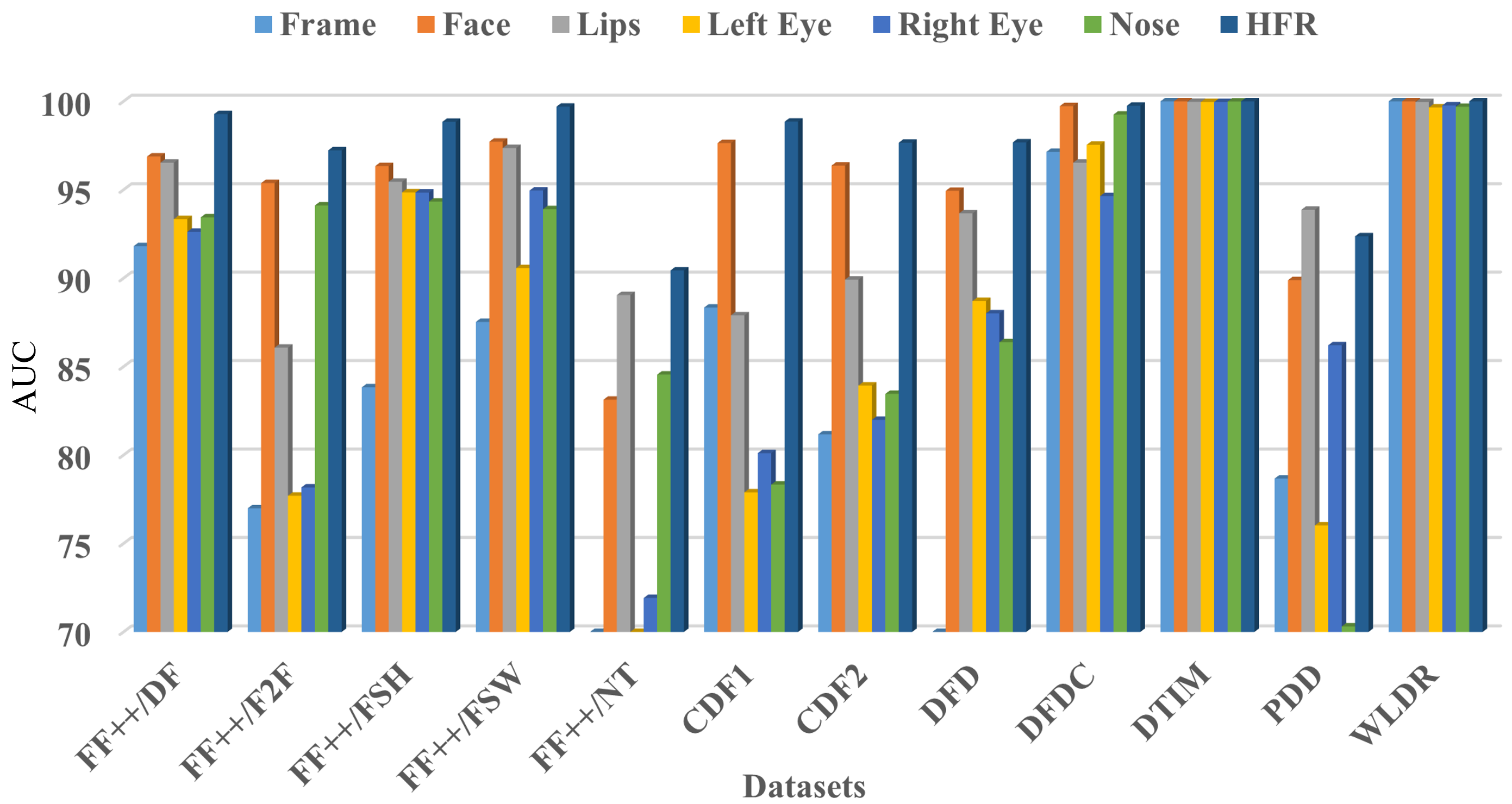}} 
    \subfloat[\label{fig:video_idx}]{\includegraphics[width=0.49\textwidth]{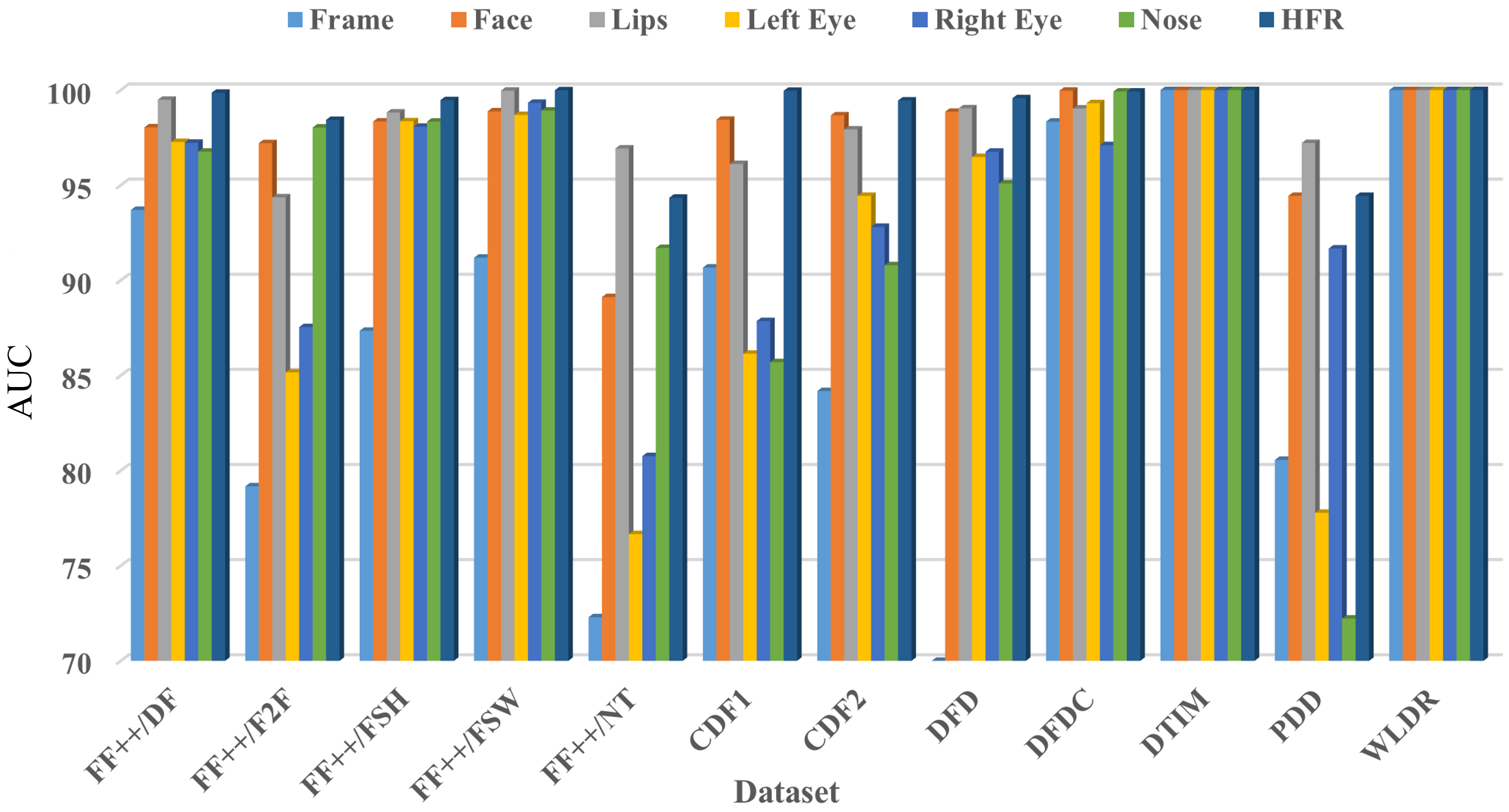}}
    \caption{Deepfake detection performance across different information levels: (a) frame-level AUC and (b) video-level AUC evaluated on eight benchmark datasets.}
    \label{fig:idx}
\end{figure*}

\begin{figure*}[!t]
    \centering
    {{\includegraphics[width=1\textwidth]{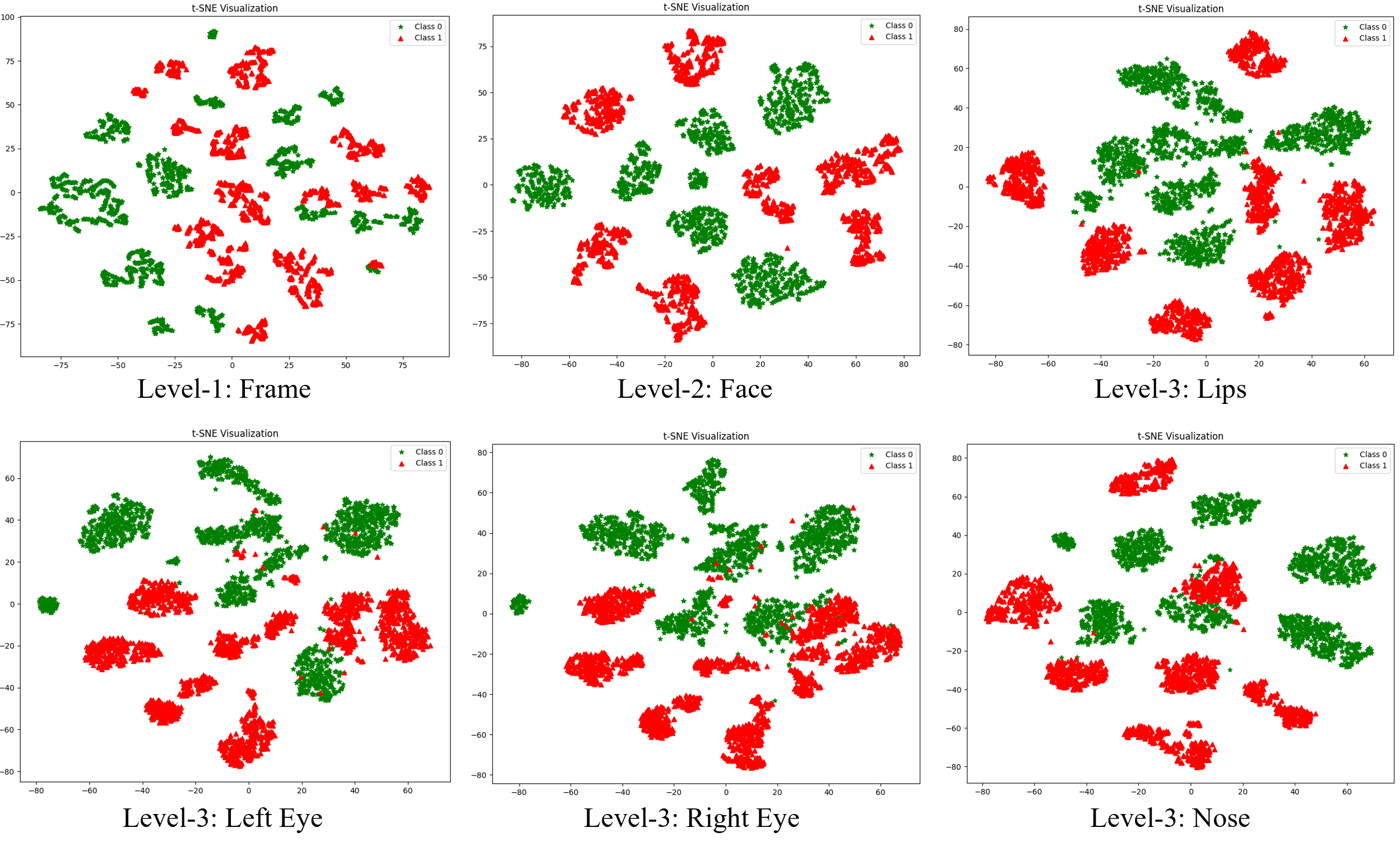}}}
    \caption{Hierarchical feature representations are extracted using a pretrained CLIP model across three levels: level-1 (frame), level-2 (face), and level-3 (lips, left eye, right eye, and nose regions).}
    \label{fig:feat}
    \vspace{-12pt}
\end{figure*}

\begin{figure*}[!t]
\centering
{{\includegraphics[width=1\textwidth]{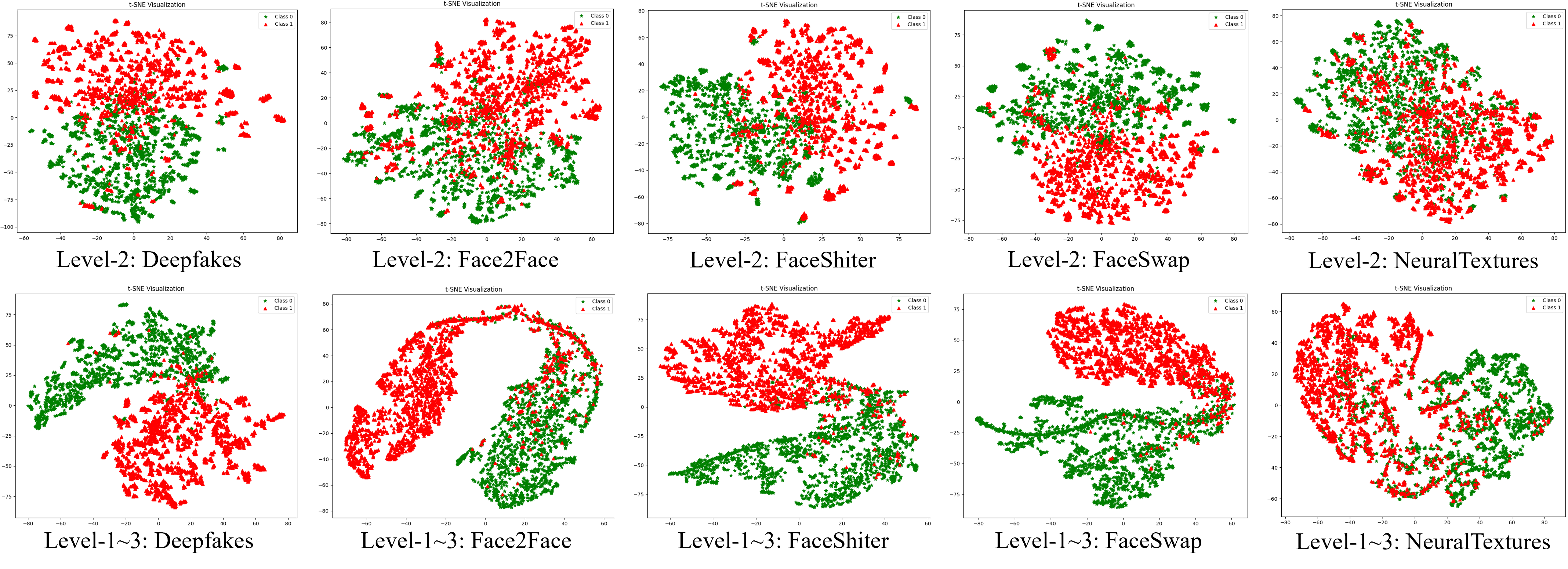}}}
\caption{t-SNE visualization of hierarchical feature embeddings after triplet learning. The top row shows embeddings using only level-2 features, which are commonly adopted in deepfake detection, while the bottom row illustrates embeddings obtained using multi-level hierarchical features from level-1 to level-3.}
\label{fig:feat_6}
\vspace{-8pt}
\end{figure*}

\begin{figure*}[!t]
    \centering
    {{\includegraphics[width=1\textwidth]{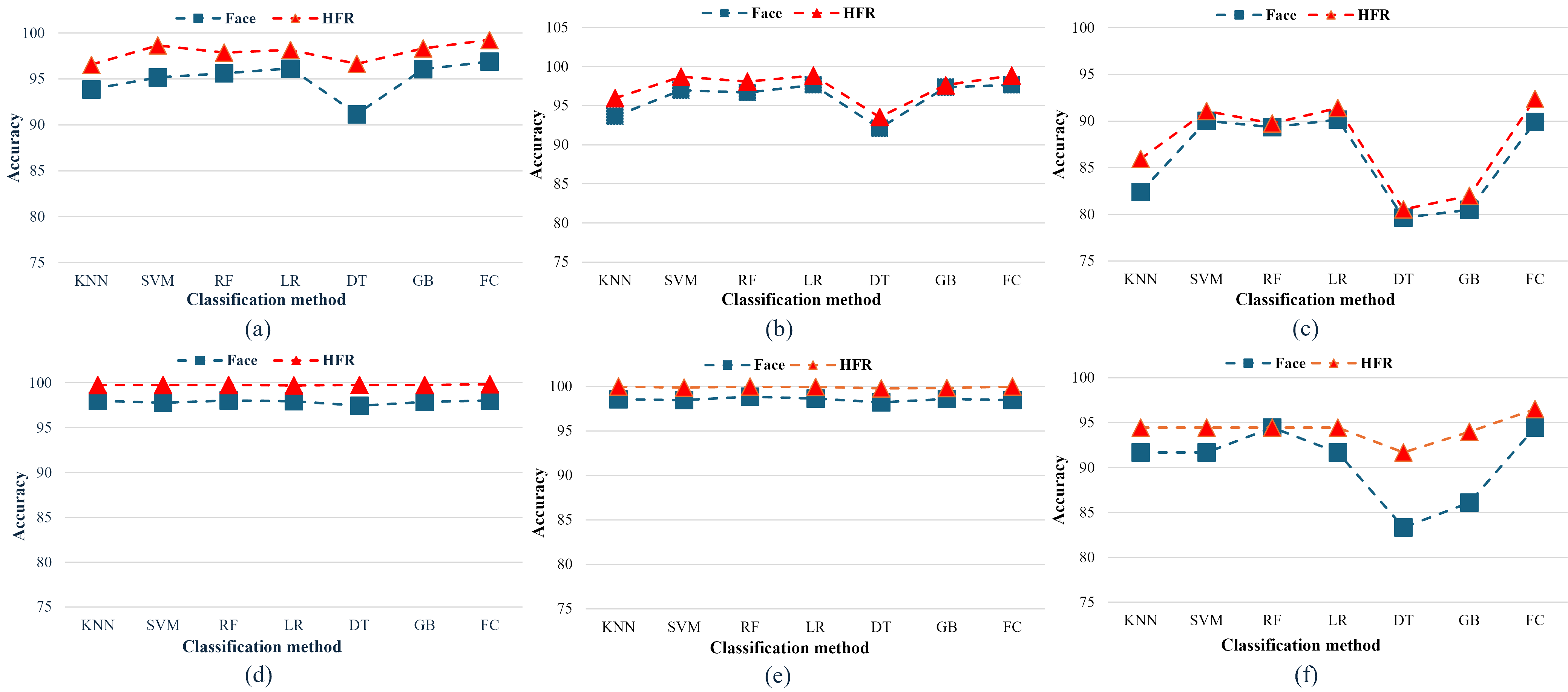}}}
    \caption{Performance comparison of different classification methods using Face and HFR features. (a)-(c) depict frame-level AUC, while (d)-(f) illustrate video-level AUC for FF++~\cite{rossler2019faceforensics++}/DF, CDF1~\cite{li2020celeb}, and PDD~\cite{sankaranarayanan2021presidential}, respectively.}
    \label{fig:ml}
    \vspace{-10pt}
\end{figure*}

\subsection{Performance Evaluations and Comparisons using Spatial Encoders}
We conducted extensive experiments with three pre-trained models across eight datasets to demonstrate the effectiveness of the proposed method. We also reported inter-dataset and inter-manipulation results to show their generalizability.

\subsubsection{Intra-Dataset Results using Spatial Encoders}
For intra-dataset evaluations, we report ACC on the FF++~\cite{rossler2019faceforensics++}, CDF1~\cite{li2020celeb}, CDF2~\cite{li2020celeb}, DFD~\cite{dolhansky2020deepfake}, DFDC~\cite{dolhansky2020deepfake}, DTIM~\cite{korshunov2018deepfakes}, PDD~\cite{sankaranarayanan2021presidential}, and WLDR~\cite{agarwal2019protecting} datasets in Table~\ref{tab:intra}. At the frame level, the proposed $HFR$ consistently outperforms face-only features across all datasets and encoders. For instance, using ViT-B16~\cite{cherti2023reproducible}, HFR improves results on FF++~\cite{rossler2019faceforensics++}/DF from 97.32\% to 99.37\%, on CDF1~\cite{li2020celeb} from 90.03\% to 93.46\%, and on PDD~\cite{sankaranarayanan2021presidential} from 78.29\% to 86.79\%. Similar improvements are observed for ViT-B32~\cite{cherti2023reproducible} and ViT-L14~\cite{cherti2023reproducible}, with HFR achieving the best performance of 99.70\% on FF++/FSW (ViT-L14~\cite{cherti2023reproducible}), highlighting its ability to leverage hierarchical features.  \\
At the video level, HFR further enhances performance, particularly for challenging datasets. With ViT-B16~\cite{cherti2023reproducible}, ACC increases from 98.66\% to 99.89\% on FF++~\cite{rossler2019faceforensics++}/DF, from 95.77\% to 98.47\% on CDF2~\cite{li2020celeb}, and from 83.33\% to 91.67\% on PDD~\cite{sankaranarayanan2021presidential}. Notably, HFR achieves perfect detection (100\% ACC) on DTIM~\cite{korshunov2018deepfakes} and WLDR~\cite{agarwal2019protecting} datasets, demonstrating robustness across different deepfake generation techniques. Overall, the results illustrate that integrating hierarchical coarse-to-fine facial features significantly improves discriminative capability, leading to superior performance over face-only features, and the gains are more pronounced at the video level, where temporal consistency can be exploited.

\subsubsection{Inter-Dataset Results using Spatial Encoders}
For inter-dataset evaluations, we train the model on FF++~\cite{rossler2019faceforensics++} and test on CDF1~\cite{li2020celeb}, CDF2~\cite{li2020celeb}, DFD~\cite{dolhansky2020deepfake}, DFDC~\cite{dolhansky2020deepfake}, DTIM~\cite{korshunov2018deepfakes}, PDD~\cite{sankaranarayanan2021presidential}, and WLDR~\cite{agarwal2019protecting}, as reported in Table~\ref{tab:inter}. Consistent with intra-dataset results, the proposed $HFR$ method substantially improves generalization performance across all benchmark datasets compared to using face-only features. For instance, at the frame level with ViT-B16~\cite{cherti2023reproducible}, HFR increases AUC on PDD~\cite{sankaranarayanan2021presidential} from 64.69\% to 81.30\% and on WLDR~\cite{agarwal2019protecting} from 77.49\% to 95.70\%. Similar improvements are observed across other encoders, with the ViT-L14~\cite{cherti2023reproducible} achieving the highest frame-level AUC of 98.78\% on WLDR~\cite{agarwal2019protecting}. At the video level, HFR also consistently boosts performance, for example, improving CDF1~\cite{li2020celeb} from 70.03\% to 76.64\% (ViT-B16~\cite{cherti2023reproducible}) and achieving perfect generalization on WLDR with 100\% (ViT-B32~\cite{cherti2023reproducible} and ViT-L14~\cite{cherti2023reproducible}). These results highlight that integrating hierarchical facial features enables the model to capture discriminative patterns that generalize well to unseen datasets, enhancing robustness against diverse deepfakes~\cite{rossler2019faceforensics++}.

\subsubsection{Performance Comparisons using Spatial Encoders}
We selected representative SoTA methods, including EfficientB4~\cite{yan2023deepfakebench}, Xception~\cite{yan2023deepfakebench}, F3Net~\cite{yan2023deepfakebench}, CORE~\cite{ni2022core}, Recce~\cite{cao2022end}, UCF~\cite{yan2023ucf}, SSL~\cite{chen2022self}, PFG~\cite{lin2024preserving}, GM-DF~\cite{lai2024gm}, SFGD~\cite{wang2023dynamic}, DSM~\cite{zhang2025dsm}, {EFFT~\cite{yan2024orthogonal}, and {D3~\cite{zheng2025d3} to compare their performance with the proposed $HFR$ method, as summarized in Table~\ref{tab:per_comparison}. For EfficientB4~\cite{yan2023deepfakebench}, Xception~\cite{yan2023deepfakebench}, and F3Net~\cite{yan2023deepfakebench}, we report the results from DeepfakeBench~\cite{yan2023deepfakebench}. Across all datasets, $HFR$ demonstrates superior generalization compared to existing methods, consistently improving detection AUC, particularly on challenging cross-dataset scenarios.  \\
The proposed $HFR$ achieves the highest average AUCs of 89.41\% on DFD~\cite{dolhansky2020deepfake}, 84.07\% on DFDC~\cite{dolhansky2020deepfake}, 95.62\% on DTIM~\cite{korshunov2018deepfakes}, 80.76\% on PDD~\cite{sankaranarayanan2021presidential}, and 100\% on WLDR~\cite{agarwal2019protecting}, surpassing all compared SoTA methods. On the FF++~\cite{rossler2019faceforensics++}, CDF1~\cite{li2020celeb}, and CDF2~\cite{li2020celeb} datasets, $HFR$ achieves the second-highest performance, closely following DSM~\cite{zhang2025dsm} and LSDA~\cite{lai2024gm}. This pattern indicates that while some specialized SoTA methods can slightly outperform $HFR$ on in-domain or simpler datasets, $HFR$ consistently maintains high performance across diverse and cross-domain datasets, demonstrating its robust generalization capability.  \\
A clear trend emerges where methods relying solely on face features tend to underperform on datasets with subtle or localized manipulations, such as PDD~\cite{sankaranarayanan2021presidential} and WLDR~\cite{agarwal2019protecting}, whereas $HFR$, leveraging hierarchical and multi-region facial features, effectively captures discriminative information across both global and local regions. This ability enables $HFR$ to outperform most SoTA methods on challenging datasets, while also maintaining competitive performance on standard benchmarks, FF++~\cite{rossler2019faceforensics++}, CDF1~\cite{li2020celeb}, and CDF2~\cite{li2020celeb}, highlighting its balanced and robust detection capabilities across diverse deepfake scenarios.

\subsubsection{Inter-Manipulation Results using Spatial Encoders}
We further compare the proposed method for inter-manipulation detection in the FF++~\cite{rossler2019faceforensics++} dataset, as shown in Table~\ref{tab:inter_comparison}. For inter-dataset comparisons, we use FF++~\cite{rossler2019faceforensics++} for training and CDF1~\cite{li2020celeb}, CDF2~\cite{li2020celeb}, DFD~\cite{dolhansky2020deepfake}, DFDC~\cite{dolhansky2020deepfake}, DTIM~\cite{korshunov2018deepfakes}, PDD~\cite{sankaranarayanan2021presidential}, and WLDR~\cite{agarwal2019protecting} for testing. We use SoTA methods of MAT~\cite{zhao2021multi}, GFF~\cite{luo2021generalizing}, DCL~\cite{sun2022dual}, and DSM~\cite{zhang2025dsm} for comparison. We reported the results of the proposed method for ViT-L14~\cite{cherti2023reproducible} model for the face alone and $HFR$. From Table~\ref{tab:inter_comparison}, it can be concluded that the proposed method is not only generalized to unseen datasets but also to unseen manipulations, surpassing the SoTA methods. It achieves the best average AUCs of 81.62\% on F2F, 93.13\% on FSW, and 79.58\% on NT manipulations when trained on DF. A similar trend is observed when training on FSW and testing on other manipulations, and achieves the best average AUCs of 92.54\% on DF, 87.46\% on F2F, and 64.56\% on NT. In most cases, the proposed method outperforms the SoTA methods.

\subsection{Performance Evaluations and Comparisons Using Temporal Encoders}
We conducted additional experiments using pre-trained MViT~\cite{fan2021multiscale}, TVC~\cite{bertasius2021space}, and ViViT~\cite{arnab2021vivit}~\cite{arnab2021vivit} models, which extract features by processing multiple frames instead of a single frame. Specifically, we used 16 frames for MViT~\cite{fan2021multiscale} and 32 frames for TVC~\cite{bertasius2021space} and ViViT~\cite{arnab2021vivit} to capture temporal dependencies across consecutive frames. The results, evaluated at both frame-level and video-level, are presented in Tables~\ref{tab:ad_intra} and~\ref{tab:ad_inter}. Here, frame-level and video-level results refer to detection performance on individual clips (e.g., 16 or 32 frames) and the entire video, respectively.

\subsubsection{Intra-Dataset Results for Temporal Encoders}
Compared to the ViT-B16~\cite{cherti2023reproducible}, ViT-B32~\cite{cherti2023reproducible}, and ViT-L14~\cite{cherti2023reproducible} models as shown in Tables~\ref{tab:intra} and ~\ref{tab:inter}, the MViT~\cite{fan2021multiscale}, TVC~\cite{bertasius2021space}, and ViViT~\cite{arnab2021vivit} models perform similarly, or sometimes better, in deepfake detection. This can be attributed to the fact that CLIP-based $ViT$ models are trained primarily on images and optimized for spatial feature extraction. These capabilities enable them to detect subtle deepfake artifacts, such as lighting inconsistencies, blending artifacts, and texture irregularities, with high effectiveness.

\subsubsection{Inter-Dataset Results for Temporal Encoders}
In contrast, MViT~\cite{fan2021multiscale}, TVC~\cite{bertasius2021space}, and ViViT~\cite{arnab2021vivit} are designed for video understanding and action recognition, which are less directly relevant for deepfake detection. However, this observation is highly dependent on the type of manipulations involved. For example, in datasets like FF++~\cite{rossler2019faceforensics++}/NT, where the manipulation type differs from other FF++~\cite{rossler2019faceforensics++} datasets, MViT~\cite{fan2021multiscale}, TVC~\cite{bertasius2021space}, and ViViT~\cite{arnab2021vivit} tend to perform better than ViT-B16~\cite{cherti2023reproducible}, ViT-B32~\cite{cherti2023reproducible}, and ViT-L14~\cite{cherti2023reproducible}. A similar trend is observed in the PDD~\cite{sankaranarayanan2021presidential} dataset, where MViT~\cite{fan2021multiscale}, TVC~\cite{bertasius2021space}, and ViViT~\cite{arnab2021vivit} outperform ViT-B16~\cite{cherti2023reproducible} and ViT-B32~\cite{cherti2023reproducible} in both frame-level and video-level deepfake detection.

\subsubsection{Inter-Manipulation Results for Temporal Encoders}
We also evaluated the inter-manipulation results for TVC~\cite{bertasius2021space}, as reported in Table~\ref{tab:ad_inter_man}. Despite this, the proposed $HFR$ method performs well, demonstrating more promising results compared to techniques that use only the face region. In all cases, the proposed method outperforms the performance achieved when using only the face region.

\subsubsection{Complexity Analysis}
We conducted a complexity analysis of the proposed $HFR$ method for ViT-B16~\cite{cherti2023reproducible}, ViT-B32~\cite{cherti2023reproducible}, ViT-L14~\cite{cherti2023reproducible}, MViT~\cite{fan2021multiscale}, TVC~\cite{bertasius2021space}, and ViViT~\cite{arnab2021vivit} models in terms of the number of parameters (in millions, M), floating point operations per second (FLOPs, in giga, G), and inference time (in seconds) for a single sample encompassing all key regions, such as the frame, face, lips, left eye, right eye, and nose, as presented in Table~\ref{tab:complex}. Compared to spatial encoders, temporal encoders require more computational resources and inference time to run the proposed method as it processes a clip of video; however, this overhead is justified given their ability to capture dynamic information. The results demonstrate that the proposed method is computationally efficient and suitable for practical deepfake detection.

\begin{table}[!t]
\centering
\caption{Performance comparisons among MLP, CA-MLP, and T.}
\begin{adjustbox}{max width=0.48\textwidth}
\begin{tabular}{lcccc}
 \midrule \midrule
 & \multicolumn{2}{c}{\textbf{Frame-level}} & \multicolumn{2}{c}{\textbf{Video-level}} \\
 \cline{2-5}
 & \textbf{FF++~\cite{rossler2019faceforensics++}/DF} & \textbf{CDF2~\cite{li2020celeb}} & \textbf{FF++~\cite{rossler2019faceforensics++}/DF} & \textbf{CDF2~\cite{li2020celeb}} \\
  \midrule
MLP & 97.32 & 96.55 & 98.26 & 97.34 \\
T+MLP & 98.98 & 97.58 & 99.17 & 98.97 \\
 \midrule
T+CA-MLP & \textbf{99.28} & \textbf{97.66} & \textbf{99.86} & \textbf{99.45} \\
 \midrule \midrule
\end{tabular}
\end{adjustbox}
\label{tab:ca_mlp}
\end{table}

\begin{table}[!t]
\centering
\caption{Choice of margin for margin loss to train the triplet model}
\begin{adjustbox}{max width=0.5\textwidth}
\begin{tabular}{ccccccccc}
\midrule \midrule
\textbf{} & \multicolumn{4}{c}{\textbf{Frame-level}} & \multicolumn{4}{c}{\textbf{Video-level}} \\
 \cline{2-9}
\textbf{} & \multicolumn{2}{c}{\textbf{FF++~\cite{rossler2019faceforensics++}}} & \multicolumn{2}{c}{\textbf{CDF2~\cite{li2020celeb}}} & \multicolumn{2}{c}{\textbf{FF++~\cite{rossler2019faceforensics++}}} & \multicolumn{2}{c}{\textbf{CDF2~\cite{li2020celeb}}} \\
\midrule
\textbf{Margin} & \textbf{Face} & \textbf{HFR} & \textbf{Face} & \textbf{HFR} & \textbf{Face} & \textbf{HFR} & \textbf{Face} & \textbf{HFR} \\
\midrule
0 & \textbf{96.88} & \textbf{99.28} & \underline{96.37} & \underline{97.66} & \textbf{98.04} & \textbf{99.86} & \textbf{98.67} & \textbf{99.45} \\
0.1 & 95.54 & 98.23 & \textbf{96.48} & \textbf{98.11} & 96.44 & 97.56 & \underline{98.55} & \underline{99.37} \\
0.2 & \underline{96.86} & \underline{99.12} & 95.67 & 97.23 & 97.12 & 98.78 & 98.01 & 98.99 \\
0.3 & 95.33 & 98.1 & 95.22 & 96.89 & \underline{97.95} & \underline{99.54} & 98.47 & 99.32 \\
\midrule \midrule
\end{tabular}
\end{adjustbox}
\label{tab:margin}
\end{table}

\begin{table}[!t]
\centering
\caption{Effectiveness of different combinations of levels}
\begin{adjustbox}{max width=0.5\textwidth}
\begin{tabular}{ccccc}
\midrule \midrule
\multirow{2}{*}{\textbf{Levels}} & \multicolumn{2}{c}{\textbf{Frame-level}} & 
\multicolumn{2}{c}{\textbf{Video-level}} \\
 \cline{2-5}
 & \textbf{FF++~\cite{rossler2019faceforensics++}/DF} & \textbf{CDF2~\cite{li2020celeb}} & \textbf{FF++~\cite{rossler2019faceforensics++}/DF} & \textbf{CDF2~\cite{li2020celeb}} \\
 \midrule
1 & 85.04 & 73.55 & 87.32 & 76.29 \\
2 & 96.88 & \multicolumn{1}{c}{96.37} & 98.04 & 98.67 \\
3 & 87.08 & 74.55 & 88.12 & 76.93 \\
\midrule
1, 2 & 96.94 & 96.88 & 98.57 & 99.01 \\
1,3 & 88.32 & 77.51 & 90.12 & 79.67 \\
2, 3 & {\underline{98.12}} & \underline{97.05} & \underline{98.96} & \underline{99.12} \\
\midrule
\textbf{1, 2, 3} & \textbf{99.28} & \multicolumn{1}{l}{\textbf{97.66}} & \textbf{99.86} & \textbf{99.45}
\\
\midrule \midrule
\end{tabular}
\end{adjustbox}
\label{tab:level}
\vspace{-15pt}
\end{table}

\subsection{Ablation Study}
This section presents an ablation study that evaluates the effectiveness of hierarchical features, learned feature representations, the CA-MLP module, triplet learning, margin optimization, individual feature performance, and various classification models.

\subsubsection{Visual Analysis of Multi-Level Feature Discriminability}
To gain deeper insight, we visualize features extracted using the ViT-L14~\cite{cherti2023reproducible} model to assess the effectiveness of each feature level, as shown in Figures~\ref{fig:frame_idx} and ~\ref{fig:video_idx} for frame-level and video-level results. These figures demonstrate that each facial region has its own strengths in detecting deepfake artifacts, depending on the type of manipulation. For example, in the FF++~\cite{rossler2019faceforensics++}/NT and PDD~\cite{sankaranarayanan2021presidential} datasets, the lips region yields better results than other feature levels. However, combining multiple levels of features provides the best performance.

\subsubsection{Visualization of Learned Feature Representations}
\label{sec:feat}
Figure~\ref{fig:feat} shows an example of extracted features for three different levels using ViT-L14~\cite{cherti2023reproducible}. We use 10,000 FF++~\cite{rossler2019faceforensics++}/DF test samples to visualize the t-SNE plot. Compared to the frame and face features, which are commonly used for deepfake detection, hierarchical features provide a more comprehensive representation that can significantly improve the overall performance in detection scenarios.\\
Figure~\ref{fig:feat_6} illustrates the hierarchical features derived from the pre-trained triplet network for the frame, face, lips, left eye, right eye, and nose. We use 10,000 FF++~\cite{rossler2019faceforensics++}/DF, FF++~\cite{rossler2019faceforensics++}/F2F, FF++~\cite{rossler2019faceforensics++}/FSH, FF++~\cite{rossler2019faceforensics++}/FSW, and FF++~\cite{rossler2019faceforensics++}/NT test samples using ViT-L14~\cite{cherti2023reproducible} model. The visualization in Figure~\ref{fig:feat_6} demonstrates that triplet learning effectively captures the relationships among features from different levels, clustering them distinctly into two groups. This highlights the network's ability to generate discriminative embeddings for real and deepfakes. 

\subsubsection{Effectiveness of CA-MLP and Triplet Learning}
We conducted an ablation study to compare the performance of MLP, MLP with T, and CA-MLP with T to demonstrate their effectiveness in capturing relationships among different facial regions, as shown in Table~\ref{tab:ca_mlp}. Notably, CA-MLP with T demonstrates superior ability to extract discriminative patterns, achieving approximately 99.28 and 97.66 at the frame level, and 99.86 and 99.45 at the video level on the FF++~\cite{rossler2019faceforensics++}/DF and CDF2~\cite{li2020celeb} datasets, respectively, outperforming both MLP and MLP with T.

\subsubsection{Triplet Loss Margin Optimization}
We experimented with a set of margin values and selected the best one to train the triplet model for optimal performance. Specifically, we evaluated margins of 0, 0.1, 0.2, and 0.3 on the FF++~\cite{rossler2019faceforensics++} and CDF2~\cite{li2020celeb} datasets, as shown in Table~\ref{tab:margin}. We reported both frame-level and video-level AUC. Notably, the proposed method achieved better results with the default margin of 0 compared to other margins (0.1, 0.2, and 0.3), except for the frame-level AUC on CDF2~\cite{li2020celeb}. Therefore, we use a margin of 0 for all subsequent experiments. This choice allows us to better exploit the relationships among embeddings from different facial regions, resulting in improved performance in intra-dataset, inter-dataset, and inter-manipulation settings.\\
While a zero margin is unconventional, it is well-suited to deepfake detection because real and fake samples already occupy naturally well-separated regions in the embedding space due to low-level artifacts. Imposing a positive margin over-constrains the optimization and, critically, hurts generalization to unseen generation methods. This is further supported by our use of foundation models, whose generic representation capability produces rich, broadly transferable embeddings that inherently reduce intra-class scatter and inter-class overlap, diminishing the need for an explicit margin to enforce separation.

\subsubsection{Impact of Hierarchical Feature on Detection Performance}
As discussed in Section~\ref{sec:feat}, we extracted three levels of features: level-1 (frame), level-2 (face), and level-3 (lips, left eye, right eye, and nose). To better evaluate the contribution of each level, we computed detection AUC for each level individually as well as for different combinations, as presented in Table~\ref{tab:level}. The results are reported using ViT-L/14 on the FF++~\cite{rossler2019faceforensics++} and CDF2~\cite{li2020celeb} datasets for both frame-level and video-level scenarios. It is evident that level-1 yields the lowest AUC due to its coarse-grained information. In contrast, combining all three levels leads to the best performance, highlighting the complementary nature of fine-grained features.

\subsubsection{Comparative Analysis of Classification Models}
After leveraging the relationship among hierarchical features through deep triplet learning, we evaluate the classification performance using various models, including traditional machine learning techniques (KNN, SVM, RF, LR, DT, GB) and deep learning models, specifically the FC model, as shown in Figure~\ref{fig:ml} for both face-alone and $HFR$ scenarios. We evaluate the models on the FF++~\cite{rossler2019faceforensics++}/DF, CDF1~\cite{li2020celeb}, and PDD~\cite{sankaranarayanan2021presidential} datasets and extract features using ViT-B32~\cite{cherti2023reproducible}. From the figure, it is evident that the FC model consistently outperforms traditional machine learning models. As discussed in previous sections, we report all results using the FC model, given its superior performance.

\begin{figure}[!t]
\vspace{-18pt}
    \centering
    \subfloat[\label{fig:ana_1}]{{\includegraphics[width=0.22\textwidth]{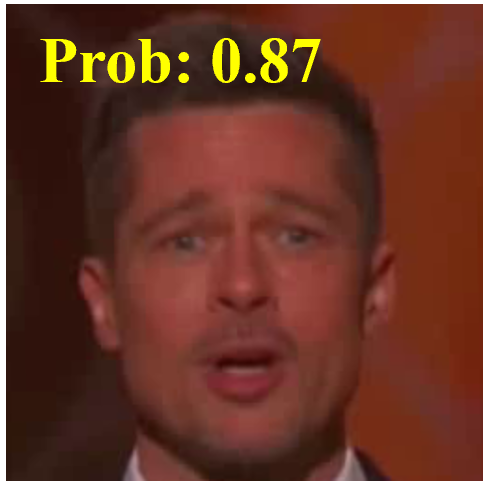}}}
    \subfloat[\label{fig:ana_2}]{{\includegraphics[width=0.22\textwidth]{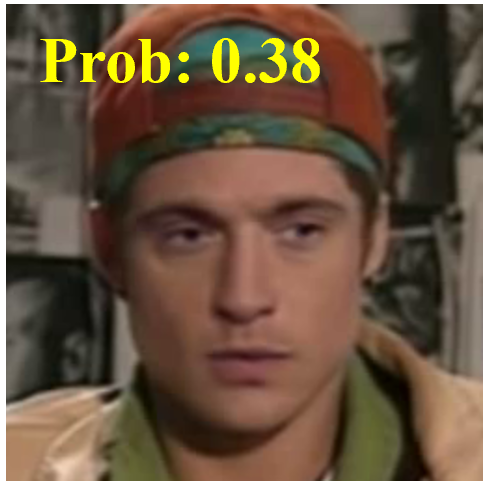}}}\\
    \subfloat[\label{fig:ana_1}]{{\includegraphics[width=0.22\textwidth]{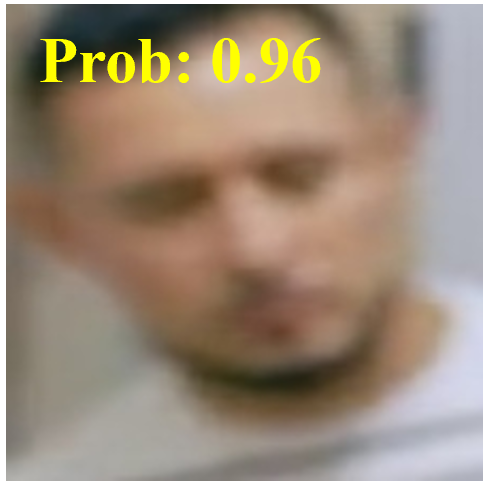}}}
    \subfloat[\label{fig:ana_2}]{{\includegraphics[width=0.22\textwidth]{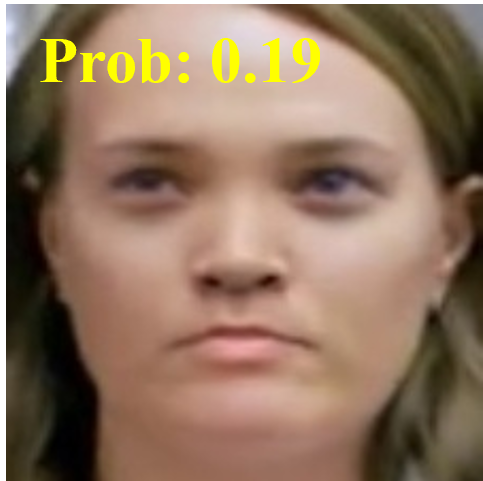}}}
    \caption{Examples of failure cases in the inter-setting evaluation. (a) CDF2~\cite{li2020celeb} real, (b) CDF2~\cite{li2020celeb} fake, (c) DFD~\cite{dolhansky2020deepfake} real, and (d) DFD~\cite{dolhansky2020deepfake} fake samples. Detection results are reported using ViT-L/14, as it achieved superior performance compared to the other evaluated detectors.}
    \label{fig:failure}
    \vspace{-15pt}
\end{figure}

\subsubsection{Quantitative and Qualitative Analysis of Failure Cases}
To better understand the detector’s limitations, we visualized four representative failure cases from the CDF2~\cite{li2020celeb} and DFD datasets, as shown in Figure~\ref{fig:failure}. Figure~\ref{fig:failure} (a) and (c) show real samples from CDF2~\cite{li2020celeb} and DFD that were incorrectly classified as fake, with high confidence scores of 0.87 and 0.96, respectively. This misclassification is mainly due to poor visual quality, including low illumination and blur, which introduces artifacts resembling synthetic traces. In contrast, Figure~\ref{fig:failure} (b) and (d) present fake samples that were incorrectly predicted as real, with confidence scores of 0.38 and 0.19. Although these samples are manipulated, they exhibit visually consistent lighting conditions and minimal blur, making them perceptually realistic and harder for the detector to distinguish from genuine content.

\section{Conclusions}
\label{sec:conclusion}
The massive use of generative models facilitates the creation of deepfakes without leaving any visual artifacts. This leads to the development of many deepfake detection methods. However, these methods suffer from generalization across datasets, while different algorithms are used to create deepfakes. In this article, we introduced a novel $HFR$ approach for enhanced and generalized deepfake detection. By leveraging coarse-to-fine-grained information from the frame, face, and key facial regions such as the lips, left eye, right eye, and nose, the proposed method effectively exploits the relationships among different facial regions using a channel-attention mechanism. This hierarchical representation enhances both intra- and inter-dataset performance, addressing the limitations of existing methods that often rely solely on the face region. Extensive evaluations on diverse datasets demonstrate the generalization capability of the proposed method.\\
Despite its promising performance, $HFR$ has some limitations, such as the computational overhead of extracting hierarchical features from multiple regions (frame, face, lips, eyes, and nose), which we plan to optimize for efficient detection. In addition, we aim to evaluate $HFR$ under adversarial conditions (e.g., filtering, compression, and generative attacks) and explore multi-modal combinations to enhance its robustness and performance further.

\newpage
\begin{IEEEbiography}
[{\includegraphics[width=1in,height=1.25in, clip, keepaspectratio]{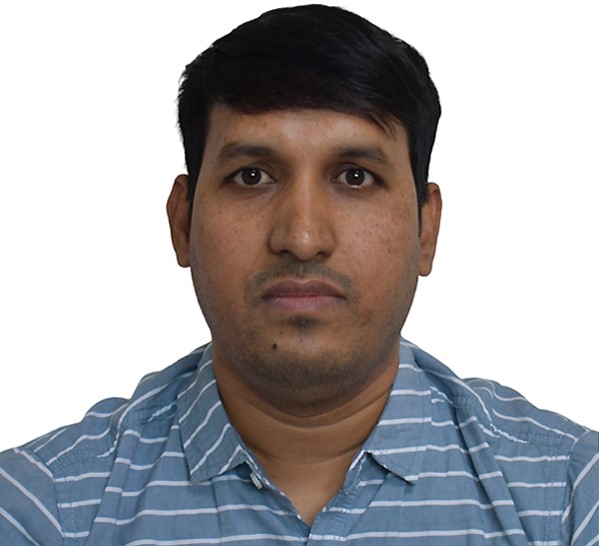}}]{Kutub Uddin}
received the B.S. degree in Computer Science and Engineering from the University of Chittagong, Chittagong, Bangladesh, in 2017. He earned his M.S. and Ph.D. degrees in Electronics and Information Engineering from Korea Aerospace University, Goyang, Korea, in 2020 and 2024, respectively. He is currently a Research Fellow in the College of Innovation and Technology at the University of Michigan, Flint, United States. His research interests include multimedia security, image, video, and audio forensics, anti-forensic, adversarial machine learning, image/video restoration and compression, vision/large-language models, human activity recognition, and 3D point cloud processing.
\end{IEEEbiography}

\begin{IEEEbiography}
[{\includegraphics[width=1in,height=1.25in, clip, keepaspectratio]{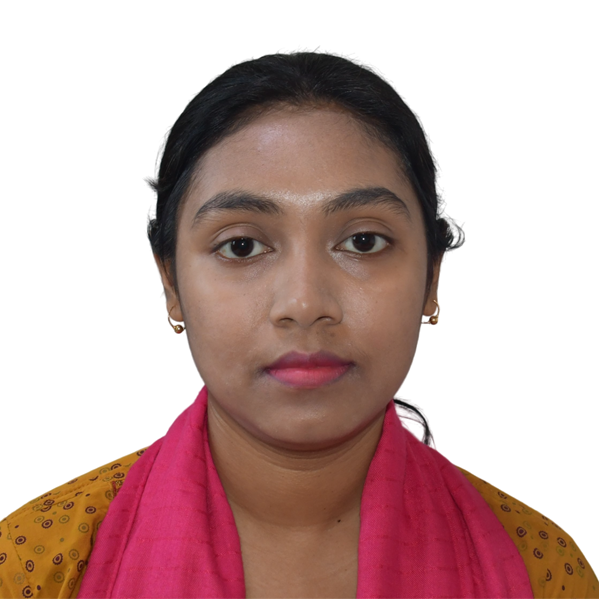}}]{Nusrat Tasnim}
received the B.S. degree in Computer Science and Engineering from the University of Chittagong, Chittagong, Bangladesh, in 2018. She earned her M.S. and Ph.D. degrees in Electronics and Information Engineering from Korea Aerospace University, Goyang, Korea, in 2020 and 2024, respectively. Her research interests include multimedia security, image, video, and audio forensics, anti-forensic, adversarial machine learning, image/video human activity recognition, and foundation models.
\end{IEEEbiography}

\begin{IEEEbiography}
[{\includegraphics[width=1in,height=1.25in, clip, keepaspectratio]{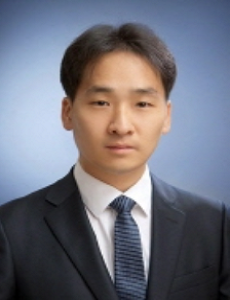}}]{Byung Tae Oh}
received the B.S. degree in Electrical Engineering at Yonsei University, Seoul, Korea, in 2003, and the M.S. and Ph.D. degrees in Electrical Engineering at the University of Southern California (USC), LA, CA in 2007 and 2009, respectively. From 2009 to 2013, he was a Research Staff with the Samsung Advanced Institute of Technology (SAIT), Samsung Electronics, Korea. Since 2013, he has been with the School of Electronics and Information Engineering, Korea Aerospace University (KAU), where he is a Professor. His research interests include image/video restoration and compression, and image/video forensics.
\end{IEEEbiography}

\vfill
\EOD

\end{document}